\documentclass[10pt,twocolumn,letterpaper]{article}

\usepackage[pagenumbers]{cvpr}

\usepackage[utf8]{inputenc}
\usepackage[T1]{fontenc}
\usepackage{url}
\usepackage{booktabs}
\usepackage{amsfonts}
\usepackage{nicefrac}
\usepackage{microtype}
\usepackage{xcolor}

\usepackage{graphicx}
\usepackage{algorithm}
\usepackage{algpseudocode}
\usepackage{amsmath,amssymb}
\usepackage{subcaption}
\usepackage{multirow}
\usepackage{wrapfig}

\newcommand{\methodfull}{Discretization Density Mixing}
\newcommand{\method}{DDM}
\newcommand{\dataR}{R}
\newcommand{\dataG}{G}
\newcommand{\dataN}{\ensuremath{\varnothing}}
\usepackage{colortbl}
\usepackage{cuted}
\usepackage{pifont}
\usepackage{array}
\usepackage{float}

\makeatletter
\patchcmd{\@maketitle}{\vspace*{12pt}}{\vspace*{-8pt}}{}{\errmessage{Title block spacing patch failed}}
\patchcmd{\@maketitle}{\vspace*{24pt}}{\vspace*{12pt}}{}{\errmessage{Title to author spacing patch failed}}
\patchcmd{\abstract}{\vspace*{12pt}}{\vspace*{6pt}}{}{\errmessage{Abstract heading spacing patch failed}}
\patchcmd{\endabstract}{\vspace*{12pt}}{\vspace*{-5pt}}{}{\errmessage{Abstract end spacing patch failed}}
\makeatother

\definecolor{cvprblue}{rgb}{0.21,0.49,0.74}
\usepackage[pagebackref,breaklinks,colorlinks,allcolors=cvprblue]{hyperref}

\def\confName{CVPR}
\def\confYear{2027}

\title{Training Denoising Models of Any Architecture Without Stored Images}

\author{Christos Koutlis \qquad Ioannis Kompatsiaris \qquad Symeon Papadopoulos\\
Information Technologies Institute, CERTH, Thessaloniki, Greece\\
{\tt\small \{ckoutlis,ikom,papadop\}@iti.gr}
}

\begin{document}
\maketitle
\begin{strip}
\centering
\includegraphics[width=\linewidth]{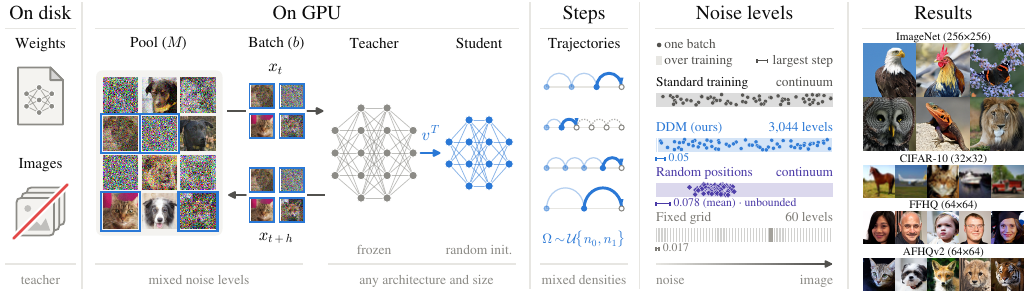}
\captionsetup{hypcap=false}
\captionof{figure}{Only the pretrained weights are stored. The teacher advances a batch of size $b$ from a pool of $M$ noisy samples by one solver step $h$ per iteration and a randomly initialized student regresses the teacher's velocity $v^T$. Every trajectory follows its own grid of $\Omega\sim\mathcal{U}\{n_0,n_1\}$ steps, thus each batch spans the noise range and, across iterations, the union of these grids approximates continuous noise levels.}
\captionsetup{hypcap=true}
\label{fig:teaser}
\end{strip}

\begin{abstract}
Can a pretrained denoising model replace the dataset used to train it? We investigate this question through teacher-supervised training, showing for the first time that continuous-time diffusion and flow models of any architecture and size can be trained from random initialization without stored images. Our students regress the teacher's velocity on a rolling pool of noisy samples, advanced along trajectories of random step counts to obtain independent, dense and approximately continuous noise levels. This saves considerable storage for large datasets and applies even when the data are unreleased or private. Our experiments include 12 teachers of 6 families and 6 datasets. Our method outperforms standard training of the same network on CIFAR-10 for students 4$\times$ smaller than the teacher, comes within 0.75 to 1.92 FID of it at the teacher's size and surpasses training on teacher-generated images and step reduction methods without teacher-based initialization. Our code is available at \url{https://github.com/mever-team/ddm}.\par
\end{abstract}

\section{Introduction}\label{sec:intro}
\providecommand{\cellstack}[1]{\begin{tabular}[t]{@{}l@{}}#1\end{tabular}}
\begin{table*}[t]
  \centering
  \small
  \caption{Related literature organized by student initialization and stored training images.}
  \label{tab:landscape}
  \begin{tabular}{@{}llll@{}}
    \toprule
    & \multicolumn{3}{c}{Stored training images} \\
    \cmidrule(l){2-4}
    Initialization & Real (\dataR) & Teacher-generated (\dataG) & None (\dataN) \\
    \midrule
    Teacher &
    \cellstack{$^{\dagger\S}$DMD2~\citep{yin2024improved}, $^\dagger$Progressive Distillation~\citep{salimans2022progressive} \\ $^\dagger$Consistency Distillation~\citep{song2023consistency}, $^\dagger$ProReflow~\citep{ke2025proreflow} \\ $^\dagger$Guided Distillation~\citep{meng2023distillation}} &
    \cellstack{$^\dagger$Reflow~\citep{liu2023flow}, $^\dagger$SCoT~\citep{fan2025scot}} &
    \cellstack{$^\dagger$BOOT~\citep{gu2023boot}, $^{\dagger\S}$SiD~\citep{zhou2024score,zhou2025score} \\ $^\dagger$$\pi$-Flow~\citep{chen2026pi}, $^\dagger$FreeFlow~\citep{tong2026flow} \\ $^\dagger$CDMD~\citep{fu2026consistent}} \\
    \midrule
    Random &
    \cellstack{$^\ddagger$Standard training~\citep{karras2022elucidating,liu2023flow,li2026back}, $^\ddagger$\citet{dockhorn2023distilling} \\ $^\dagger$Consistency Training~\citep{song2023consistency}, $^\dagger$Shortcut Models~\citep{frans2025one} \\ $^\dagger$MeanFlow~\citep{geng2025mean}} &
    \cellstack{$^\dagger$SlimFlow~\citep{zhu2024slimflow} \\ $^\ddagger$SlimFlow (2-RF)~\citep{zhu2024slimflow} \\ $^\ddagger$Random Conditioning~\citep{kim2025random}} &
    \cellstack{$^\ddagger$DKDM~\citep{xiang2025dkdm}, $^\ddagger$\textbf{\method\ (ours)}} \\
    \bottomrule
    \multicolumn{4}{@{}l}{\footnotesize $^\dagger$Step reduction method. $^\ddagger$Multi-step denoising model. $^\S$Trained by us from random initialization without stored images.} \\
  \end{tabular}
\end{table*}

Denoising models learn to restore noise-corrupted images, requiring a stored dataset whose size can reach hundreds of terabytes \cite{schuhmann2022laion}. Training therefore presupposes storage capacity and becomes impossible when the data are unreleased~\cite{esser2024scaling} or private~\cite{dockhorn2023differentially}. Moreover, the pretrained model often differs from the one desired, as deployment requires smaller networks and research new architectures. Nevertheless, it encodes the data distribution, and this raises the question whether a pretrained model can replace the dataset and be used to train a new denoising model of any architecture and size from random initialization, storing only its weights.

Tab.~\ref{tab:landscape} organizes related work by the two resources our framework excludes, teacher-based initialization and stored images, either real (\dataR) or teacher-generated (\dataG). The former transfers pretrained weights but restricts the student to the teacher's architecture and size, and the latter bounds the number of training images by the available storage. Training on stored teacher-generated images is often termed ``data-free''~\citep{kim2025random}, while we pose the stricter constraint of training \emph{without stored images} (\dataN). Step reduction~\citep{song2023consistency,yin2024improved,zhou2024score} depends on at least one of these resources, since a few-step student targets the endpoint of a long trajectory segment, which is either stored or approximated by a teacher-initialized student. Most multi-step methods~\citep{karras2022elucidating,dockhorn2023distilling,zhu2024slimflow,kim2025random} start from random initialization but require stored training images, as their objective forms noisy inputs from clean images. In the absence of both resources \cite{xiang2025dkdm}, supervision at intermediate trajectory states is required, yielding multi-step models.

DKDM~\citep{xiang2025dkdm} first demonstrated that this regime is attainable, training randomly initialized diffusion models of a different architecture without stored images. Its formulation, however, is restricted to early discrete-time Denoising Diffusion Probabilistic Models (DDPM)~\citep{ho2020denoising}, whose objective is defined on a finite set of timesteps that the teacher's denoising chain visits by construction. A continuous-time method can still use such teachers by rounding each noise level to the nearest of their timesteps~\citep{karras2022elucidating}, but then loses the continuum it is designed for and replaces their stochastic denoising chain with deterministic trajectories. Conversely, continuous-time diffusion and flow teachers introduce a harder problem since their objective is defined over a continuum of noise levels, whereas an unrolled teacher visits only its solver grid.

Standard training~\citep{karras2022elucidating,liu2023flow,li2026back} avoids this problem, as it draws the noise level of each sample \textit{independently} from a \textit{continuous distribution}. Unrolling the teacher on a fixed grid provides neither property, as trajectories initialized together keep all samples of a batch at one noise level, and the student is supervised only at the $n$ levels of the grid, which sparsely cover the continuum. A finer fixed grid increases their \textit{density}, but within a fixed training budget it markedly reduces the number of unrolled trajectories. Alternatively, randomly drawing the grid positions of each trajectory yields a continuum of noise levels, yet keeps batch noise levels concentrated and produces occasional large solver steps.

We address these limitations with \methodfull\ (\method), which draws the number of solver steps of every trajectory at random, $\Omega\sim\mathcal{U}\{n_0,n_1\}$. As trajectories of different lengths end at different iterations, the noise levels within a batch become independent. The union of the grids of all trajectories forms a dense set of noise levels that approximates a continuum, and the maximum solver step is $1/n_0$. The student is trained to match the teacher's velocity along trajectories generated online, of which only the latest state is kept in GPU memory, and no training image is stored on disk. For instance, the teacher weights are 137$\times$ smaller than the training images of ImageNet at $256\times256$. Teacher and student interact through their velocity fields, thus the student is unconstrained in architecture and size and initialized at random. \method\ outperforms standard training of the same network on CIFAR-10 for students 4$\times$ smaller than the teacher, by 0.38 FID, by 0.18 with class labels and by 1.37 with Rectified Flow. At the teacher's size, it attains 2.54 FID on class-conditional CIFAR-10 and 5.58 on ImageNet at $256\times256$, within 0.75 and 1.92 FID of standard training, respectively, and outperforms SlimFlow (2-RF)~\citep{zhu2024slimflow}, trained on stored teacher-generated images, by 0.13 to 0.80 FID. Without teacher-based initialization or stored images, step reduction methods DMD2~\citep{yin2024improved} and SiD~\citep{zhou2024score} remain above 83 FID, while \method\ attains 3.08 to 4.88. To compare directly with DKDM~\citep{xiang2025dkdm}, \method\ uses its discrete-time teachers despite the disadvantages and outperforms it in three of five settings, by up to 2.43 FID. Our contributions:
\begin{itemize}
    \item We propose the first method that trains continuous-time denoising models of any architecture and size from random initialization without stored training images.
    \item We identify independent, continuous and dense noise levels as three properties of standard training and introduce \method, which approximates all three by construction.
    \item  With step-count bounds selected once on CIFAR-10, \method\ trains students from 12 teachers of 6 families on 6 datasets in pixel and latent space, without further tuning.
    \item \method\ outperforms standard training of the same network on CIFAR-10 for students 4$\times$ smaller than the teacher, comes within 0.75 to 1.92 FID of it at the teacher's size and surpasses training on teacher-generated samples and step reduction methods without teacher-based initialization.
\end{itemize}

\section{Related work}

\textbf{Teacher-based initialization.} Most methods that learn from a pretrained diffusion or flow model reduce its sampling steps and start the student from the teacher's weights. Progressive Distillation~\citep{salimans2022progressive}, Consistency Distillation~\citep{song2023consistency}, Guided Distillation~\citep{meng2023distillation} and ProReflow~\citep{ke2025proreflow} noise real images, and DMD2~\citep{yin2024improved} trains a discriminator on real images alongside distribution matching. Reflow~\citep{liu2023flow} and SCoT~\citep{fan2025scot} instead train on noise--image pairs precomputed by the teacher. BOOT~\citep{gu2023boot}, SiD~\citep{zhou2024score,zhou2025score}, $\pi$-Flow~\citep{chen2026pi}, FreeFlow~\citep{tong2026flow} and CDMD~\citep{fu2026consistent} store no images, but supervise the student through its own predictions, which requires this initialization. Hence, all of them inherit the teacher's architecture and size. To test them in our setting, we train DMD2, without its GAN loss and real data, and SiD from random initialization.

\textbf{Random initialization with stored images.} Most methods that train from random initialization rely on stored images. Standard training of diffusion and flow models~\citep{karras2022elucidating,liu2023flow,li2026back} noises real images at every iteration, and so do few-step models trained from scratch, such as Consistency Training~\citep{song2023consistency}, Shortcut Models~\citep{frans2025one} and MeanFlow~\citep{geng2025mean}. \citet{dockhorn2023distilling} train students on real images to match the teacher's denoiser. SlimFlow~\citep{zhu2024slimflow}, whose one-step models start from its multi-step 2-RF models, and Random Conditioning~\citep{kim2025random} avoid real data, but train on hundreds of thousands of stored teacher-generated images. In every case, storage capacity bounds the number of training images. We compare against SlimFlow (2-RF), the closest to our setting, as it trains from random initialization without real data, yet on stored teacher-generated noise--image pairs.

\textbf{Random initialization without stored images.} To our knowledge, DKDM~\citep{xiang2025dkdm} is the only prior method in this setting, with students of any architecture. It replaces the dataset with the reverse process of a DDPM teacher, which denoises Gaussian samples by one step per training iteration, while the student matches the teacher's noise prediction at each visited timestep. Since this chain visits only the 1,000 discrete timesteps on which the objective of DDPM~\citep{ho2020denoising} is defined, DKDM is restricted to discrete-time teachers. \method\ addresses continuous-time training, where students must be supervised over a continuum of noise levels. Applied to discrete-time teachers, our method is disadvantaged by construction, as it rounds every noise level to one of these timesteps and follows deterministic sampling trajectories, whereas DKDM natively follows the teacher's stochastic denoising chain. Despite this disadvantage, we compare against DKDM on its discrete-time teachers.

\begin{algorithm}[t]
\caption{Training without stored images with \methodfull\ (\method). \textsc{Schedule} maps a trajectory step to the teacher's noise level and \textsc{TeacherStep} takes one teacher solver step, returning its velocity (cf. App.~\ref{appendix:teacher_step}). Conditions $\mathbf{c}\sim p_c$ apply to conditional models.}
\label{alg:teacher_sampling}
\begin{algorithmic}[1]
\Require teacher $v^T$, student $v^S$ with weights $\phi$, step-count bounds $n_0\le n_1$, batch size $b$, pool size $M\ge b$, iterations $K$, EMA decay $d$
\State $\boldsymbol{\xi}\gets\mathbf{0}\in\mathbb{N}^M,\ \boldsymbol{\omega}\sim\mathcal{U}\{n_0,n_1\}^M$
\State $\mathbf{\tilde{x}}\sim p_0^M,\ \mathbf{c}\sim p_c^M,\ \bar{\phi}\gets\phi$
\For{$k=1,\dots,K$}
  \State $\mathcal{J}\gets\textsc{Sample}(M,b)$
  \State $\mathcal{C}\gets\{i\in\mathcal{J}:\xi_i=\omega_i\}$
  \State $\boldsymbol{\xi}_\mathcal{C}\gets\mathbf{0},\ \boldsymbol{\omega}_\mathcal{C}\sim\mathcal{U}\{n_0,n_1\}^{|\mathcal{C}|}$
  \State $\mathbf{\tilde{x}}_\mathcal{C}\sim p_0^{|\mathcal{C}|},\ \mathbf{c}_\mathcal{C}\sim p_c^{|\mathcal{C}|}$
  \State $\mathbf{t}\gets\textsc{Schedule}(\boldsymbol{\xi}_\mathcal{J},\boldsymbol{\omega}_\mathcal{J})$
  \State $\mathbf{t}'\gets\textsc{Schedule}(\boldsymbol{\xi}_\mathcal{J}+\mathbf{1},\boldsymbol{\omega}_\mathcal{J})$
  \State $(\mathbf{\tilde{x}}',\mathbf{v})\gets\textsc{TeacherStep}(v^T;\mathbf{\tilde{x}}_\mathcal{J},\mathbf{t},\mathbf{t}',\mathbf{c}_\mathcal{J})$
  \State $\mathcal{L}\gets\frac{1}{b}\|v^S(\phi;\mathbf{\tilde{x}}_\mathcal{J},\mathbf{t},\mathbf{c}_\mathcal{J})-\mathbf{v}\|_2^2$
  \State $\phi\gets\textsc{OptimizerStep}(\phi,\nabla_\phi\mathcal{L})$
  \State $\bar{\phi}\gets d\,\bar{\phi}+(1-d)\,\phi$
  \State $\mathbf{\tilde{x}}_\mathcal{J}\gets\mathbf{\tilde{x}}',\ \boldsymbol{\xi}_\mathcal{J}\gets\boldsymbol{\xi}_\mathcal{J}+\mathbf{1}$
\EndFor
\State \Return $\bar{\phi}$
\end{algorithmic}
\end{algorithm}

\section{Method}
\method's detailed implementation is presented in Alg.~\ref{alg:teacher_sampling}.

\subsection{Background}\label{subsec:background}
Denoising models are trained on a Gaussian path $\mathbf{x}_t=\alpha_t\mathbf{x}+\beta_t\mathbf{z}$ between data $\mathbf{x}$, from a training set $\mathcal{D}$ in pixel or latent space~\citep{rombach2022high}, and noise $\mathbf{z}\sim\mathcal{N}(\mathbf{0},\mathbf{I})$. A network $\mathbf{f}_\theta$ regresses the data, the noise or the velocity $\dot{\mathbf{x}}_t=\dot{\alpha}_t\mathbf{x}+\dot{\beta}_t\mathbf{z}$, by solving
\begin{equation}\label{eq:objective-data-driven}
    \min_\theta\,\mathbb{E}_{t,\mathbf{x},\mathbf{z}}\big[\lambda_t\|\mathbf{f}_\theta(\mathbf{x}_t,t)-\mathbf{f}_t\|^2\big],\quad\mathbf{f}_t\in\{\mathbf{x},\mathbf{z},\dot{\mathbf{x}}_t\},
\end{equation}
with $t\sim p(t)$ and a weighting $\lambda_t$. The prediction of $\mathbf{f}_\theta$ thereby estimates the posterior mean of its target, the noise~\citep{ho2020denoising}, the data~\citep{karras2022elucidating,li2026back} or the velocity~\citep{lipman2023flow,liu2023flow}, with predictions $\mathbf{z}_\theta$, $\mathbf{x}_\theta$ and $\mathbf{v}_\theta$, respectively. As $\mathbf{x}_t$ is a linear combination of $\mathbf{x}$ and $\mathbf{z}$, the three posterior means determine one another, and in particular the velocity
\begin{equation}\label{eq:velocity}
    v(\theta;\mathbf{x}_t,t)=
    \begin{cases}
        \mathbf{v}_\theta & \text{if }\mathbf{f}_t=\dot{\mathbf{x}}_t,\\
        (\dot{\beta}_t\mathbf{x}_t+\gamma_t\mathbf{x}_\theta)/\beta_t & \text{if }\mathbf{f}_t=\mathbf{x},\\
        (\dot{\alpha}_t\mathbf{x}_t-\gamma_t\mathbf{z}_\theta)/\alpha_t & \text{if }\mathbf{f}_t=\mathbf{z},
    \end{cases}
\end{equation}
with $\gamma_t=\dot{\alpha}_t\beta_t-\alpha_t\dot{\beta}_t$. This velocity transports $p_0$ to the data distribution~\citep{song2020score,lipman2023flow}, and solving the ordinary differential equation (ODE) $d\mathbf{x}_t/dt=v(\theta;\mathbf{x}_t,t)$ generates samples $\mathbf{\hat{x}}\sim p_1^\theta$. Hereafter, teacher and student interact through their velocities $v^T(\theta;\mathbf{x}_t,t)$ and $v^S(\phi;\mathbf{x}_t,t)$, which any prediction target yields by Eq.~\eqref{eq:velocity} and the teacher's solver integrates.

\subsection{Trajectory unrolling and velocity regression}
\begin{figure*}[t]
    \centering
    \includegraphics[width=\textwidth]{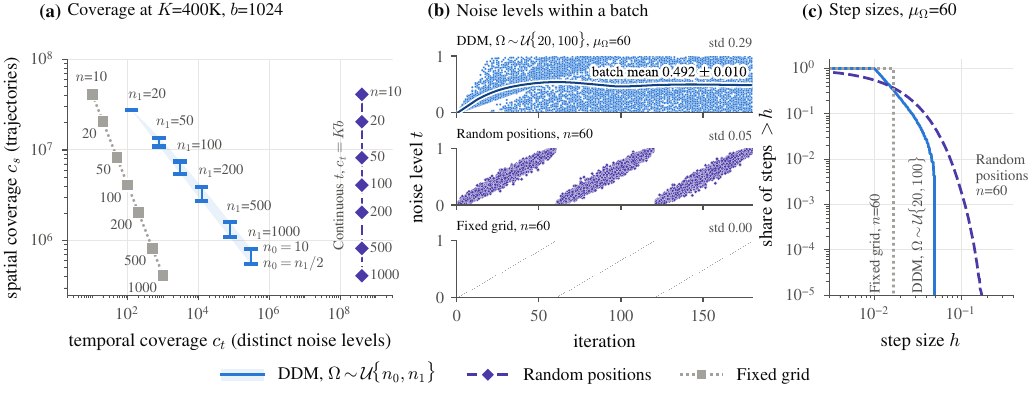}
    \caption{Coverage, batch composition and step sizes of \method, random positions and a fixed grid. (a) Spatial coverage $c_s$ against temporal coverage $c_t$ at $K=400$K and $b=1024$, with $n$ from 10 to 1000 for the fixed grid and random positions. Each \method\ bar spans $n_0$ from 10 (top) to $n_1/2$ (bottom) at one $n_1$, and the band covers $n_1$ from 20 to 1000. (b) Noise levels of one batch in the first 180 iterations, labelled with their steady-state standard deviation. (c) Share of steps larger than $h$ in a simulated run of $10^6$ steps, with largest step bounded by 0.05 for \method, 0.017 for the fixed grid and unbounded for random positions, with mean 0.078, maximum 0.21 and standard deviation 0.019.}
    \label{fig:coverage}
\end{figure*}
Without the training set $\mathcal{D}$, the inputs $\mathbf{x}_t$ of Eq.~\eqref{eq:objective-data-driven} cannot be formed. The direct remedy replaces the data by teacher-generated samples $\mathbf{\hat{x}}\sim p_1^\theta$, which gives
\begin{equation}\label{eq:objective-data-free-heavy}
    \min_\phi\,\mathbb{E}_{t,\mathbf{\hat{x}},\mathbf{z}}\big[\|v^S(\phi;\mathbf{\hat{x}}_t,t)-(\dot{\alpha}_t\mathbf{\hat{x}}+\dot{\beta}_t\mathbf{z})\|^2\big],
\end{equation}
with $\mathbf{\hat{x}}_t=\alpha_t\mathbf{\hat{x}}+\beta_t\mathbf{z}$. Each training iteration then costs a full ODE solve of $n$ teacher evaluations, whose intermediate states are discarded. Generating these samples on the fly is therefore impractical, and precomputing them bounds their number by the available GPU memory or disk storage.

We instead unroll the teacher's trajectories one solver step per iteration and supervise the student with the teacher's velocity at every generated state, thereby wasting no teacher evaluation. A trajectory starts from $\mathbf{\tilde{x}}\sim p_0$ and follows a grid of $n$ steps. After $\xi\in\{0,\dots,n-1\}$ steps, its state has the normalized position $\xi/n\in[0,1)$, which the teacher's schedule maps to the noise level $t$ (App.~\ref{appendix:teacher_step}). Between consecutive grid levels $t$ and $t+h$, the state advances as
\begin{equation}\label{eq:update}
    \mathbf{\tilde{x}}_{t+h}=\mathbf{\tilde{x}}_t+h\,\mathbf{v}^T_t,
\end{equation}
where $\mathbf{v}^T_t$ is the velocity of the teacher's Euler or Heun step~\citep{hairer1993solving}. The same velocity is the student's target in
\begin{equation}\label{eq:objective}
    \min_\phi\,\mathbb{E}\big[\|v^S(\phi;\mathbf{\tilde{x}}_t,t)-\mathbf{v}^T_t\|^2\big],
\end{equation}
where $t$ ranges over the noise levels of the grid and $\mathbf{\tilde{x}}_t$ over the states of the trajectories at noise level $t$. This procedure requires no disk storage, since only the latest state of every trajectory is kept, and it is model-agnostic, since teacher and student interact only through their velocity predictions.

\subsection{\methodfull\ (\method)}\label{subsec:ddm}\label{subsec:progressive_sampling}
On a fixed grid of $n$ steps, Eq.~\eqref{eq:objective} supervises the student at only $n$ noise levels. A finer grid increases them at the cost of fewer trajectories (Fig.~\ref{fig:coverage}a), while trajectories initialized together keep each batch at one noise level (Fig.~\ref{fig:coverage}b).

\method\ removes both restrictions by randomly assigning each trajectory a number of steps $\Omega\sim\mathcal{U}\{n_0,n_1\}$ drawn uniformly between the step-count bounds $n_0$ and $n_1$. The trajectories then advance along the teacher's schedule in $\omega$ steps and the student is supervised on the union of the grids of all step counts, whose density of distinct noise levels is
\begin{equation}\label{eq:levels}
    c_t=\left|\bigcup_{\omega=n_0}^{n_1}\left\{\frac{\xi}{\omega}\;\middle|\;\xi=0,\dots,\omega-1\right\}\right|.
\end{equation}
The latest state of every trajectory is kept in a noisy sample pool of $M\ge b$ samples, and at each iteration a random subset of size $b$ forms the batch that is advanced by one solver step. Every trajectory that completes its grid restarts from a new sample of $p_0$ with a new step count. The starts of the successive trajectories of a sample thus form a renewal process~\citep{ross1996stochastic} with i.i.d. step counts $\Omega$, and over the course of training the normalized positions $\tau=\xi/\Omega$ follow (App.~\ref{appendix:levels})
\begin{equation}\label{eq:position_cdf}
    P(\tau\le t)=\frac{1}{\mu_\Omega}\sum_\omega P(\Omega=\omega)\cdot\big(\lfloor\omega t\rfloor+1\big),
\end{equation}
where $\mu_\Omega$ is the mean of $\Omega$. Since trajectories of different lengths end at different iterations, Eq.~\eqref{eq:position_cdf} also describes the distribution within each batch, even when $M=b$.

\textbf{Corollary} (App.~\ref{appendix:levels}). \textit{For \method\ with $\Omega\sim\mathcal{U}\{n_0,n_1\}$, $n_1>n_0$ and $M=b$, the positions within a batch become independent samples of a distribution that differs from $\mathcal{U}(0,1)$ by at most $1/\mu_\Omega=2/(n_0+n_1)$ at every $t$.}

\subsection{Discretization schemes}\label{subsec:schemes}
We compare \method\ with two discretization schemes of $n$ steps, the fixed grid, whose steps are equally spaced, and random positions, drawn uniformly at random over a continuum. Without a pool, $M=b$, only \method\ yields, by construction, independent noise levels within a batch (Fig.~\ref{fig:coverage}b), with a standard deviation of 0.29, as for $\mathcal{U}(0,1)$, against 0.05 for random positions and 0 for the fixed grid. A pool of $M>b$ samples renders the noise levels of the other two schemes independent as well, and since the positions of a fixed grid deviate from uniform by at most $1/n$, all three schemes provide independent and at least approximately uniform noise levels. They differ, however, in the density of noise levels, which the fixed grid limits to $n$, and in the size of solver steps, which random positions leave uncontrolled, with individual steps several times larger than $1/n$ (Fig.~\ref{fig:coverage}c). \method\ resolves both, as the union of its grids yields a dense set of noise levels, Eq.~\eqref{eq:levels}, and none of its steps exceeds $1/n_0$.

\subsection{Velocity field coverage}\label{subsec:coverage}
The choice of the step-count bounds $n_0$ and $n_1$ governs a trade-off between spatial and temporal coverage of the teacher's velocity field. Over $K$ iterations, the trajectories form a renewal process~\citep[Definition 3.1.1, p. 98]{ross1996stochastic} with i.i.d. lengths $\Omega\sim\mathcal{U}\{n_0,n_1\}$ of mean $\mu_\Omega=(n_0+n_1)/2$, whose number has the asymptotic mean $Kb/\mu_\Omega$~\citep[Theorem 3.3.5, p. 108]{ross1996stochastic}. We therefore define the spatial coverage as this expected number of trajectories, $c_s=Kb/\mu_\Omega=2Kb/(n_0+n_1)$, and the temporal coverage $c_t$ as the number of distinct noise levels of Eq.~\eqref{eq:levels}. Fig.~\ref{fig:coverage}a places the three schemes in the $(c_t,c_s)$ plane. As $c_s=Kb/\mu_\Omega$ for all of them, they differ in temporal coverage, that is, in the density of their noise levels, at equal mean step count. A fixed grid of $n$ steps supervises only $c_t=n$ noise levels and lies on the line $c_sc_t=Kb$. Random positions reach $c_t=Kb$, as no noise level repeats. At the spatial coverage of a fixed grid with the same mean step count, the temporal coverage of \method\ grows as $n_1^2$ for $n_0\le n_1/2$, placing it above the line $c_sc_t=Kb$, between the fixed grid and random positions.

\section{Experimental setup}\label{sec:setup}
\textbf{Teacher and student architectures.} Our teachers are the released checkpoints of EDM~\citep{karras2022elucidating} and Rectified Flow~\citep{liu2023flow}, both DDPM++ networks~\citep{song2020score}, on CIFAR-10~\citep{krizhevsky2009learning} ($32\times32$), FFHQ~\citep{karras2019style} and AFHQv2~\citep{choi2020stargan,karras2021alias} ($64\times64$), of JiT-L/16~\citep{li2026back} on ImageNet~\citep{deng2009imagenet} ($256\times256$), of the ADM models of DDPM-IP~\citep{ning2023input} on CIFAR-10, ImageNet ($32\times32$) and CelebA~\citep{liu2015deep} ($64\times64$), of the latent LDM-VQ-4 models~\citep{rombach2022high} on CelebA-HQ~\citep{karras2018progressive} and FFHQ ($256\times256$), and of U-ViT-S/2~\citep{bao2023all} on CIFAR-10. Each student has the architecture of its teacher, at the teacher's size or two to five times smaller, except in the cross-architecture experiments, where DDPM++ and DiT~\citep{peebles2023scalable} students learn from the EDM and U-ViT teachers.

\textbf{Training details.} Students are randomly initialized and trained with a pool of $M=3b$ samples, which the teacher advances with Heun's solver at step-count bounds $\{n_0,n_1\}=\{20,100\}$ for the diffusion and $\{10,50\}$ for the flow teachers. These bounds are selected once on CIFAR-10 and reused without further tuning. We use Adam~\citep{kingma2015adam} with $(\beta_1,\beta_2)=(0.9,0.999)$, a learning rate of $2\times10^{-4}$ after 5K linear warmup iterations, an exponential moving average (EMA) with decay 0.9999, a batch size of $b=1024$ and $K=400$K iterations, except for $b=2048$ on ImageNet, $b=512$ on CelebA and in latent space, and $K=500$K on ImageNet and in latent space. The JiT student uses $M=4b$, and its teacher is unrolled with classifier-free guidance~\citep{ho2022classifier} at scale 3.0 for $t\in[0.1,1]$. The ablations train the 14.0M EDM and 15.5M Rectified Flow students on CIFAR-10 for 80K and 100K iterations. 
App.~\ref{appendix:training} lists all settings.

\textbf{Competing methods.} No prior method trains continuous-time students of any architecture from random initialization without stored images. We therefore compare against methods with an inherent advantage, namely standard training of the same architecture and size on real images, SlimFlow's 2-RF models~\citep{zhu2024slimflow} trained on stored teacher-generated images, and DKDM~\citep{xiang2025dkdm}, the closest method to ours, on the discrete-time ADM and LDM teachers it is designed for. We further train DMD2~\citep{yin2024improved} (without GAN loss) and SiD~\citep{zhou2024score} from random initialization without stored images.

\textbf{Evaluation protocol.} Unless otherwise stated, we report the Fr\'echet Inception Distance (FID)~\citep{heusel2017gans} on 50K samples, against reference statistics of each teacher's training set (App.~\ref{appendix:fid_stats}). Students use their teacher's sampler, whose cost we report in neural function evaluations (NFE), namely Heun's solver~\citep{karras2022elucidating} with 18 steps (35 NFE) at $32\times32$ and 40 steps (79 NFE) at $64\times64$ for EDM, RK45 for Rectified Flow, and Heun's solver with classifier-free guidance (CFG) at 198 NFE for JiT. For the ADM teachers, whose ancestral sampler requires a learned variance absent from our students, we use EDM's stochastic Heun sampler with 50 steps and $S_\text{churn}=10$, the best of $\{5,10,40\}$ for our students (99 NFE), and for the LDM teachers DDIM~\citep{song2021denoising} with 50 steps.

\section{Results}\label{sec:results}
\subsection{Comparison with standard training}\label{subsec:data_based}
\begin{table}[t]
  \centering
  \footnotesize
  \setlength{\tabcolsep}{3.4pt}
  \caption{\method\ vs. standard training in FID$\downarrow$. ``\dataR'' denotes stored real images, \dataN\ none, ``$\Delta$'' the difference from standard training and \textbf{bold} the settings in which \method\ outperforms standard training.}
  \label{tab:students_vs_teachers}
  \begin{tabular}{rrccr}
\toprule
Params & NFE & Standard training (\dataR) & \method\ (\dataN) & $\Delta$ \\
\midrule
\addlinespace
\multicolumn{5}{l}{\textbf{\boldmath EDM~\citep{karras2022elucidating}, CIFAR-10 ($32\times32$) $\cdot$ Heun}} \\
14.0M & 35 & 3.94 & 3.56 & \textbf{\boldmath$-$0.38} \\
55.7M & 35 & 1.97 & 2.79 & +0.82 \\
\midrule
\addlinespace
\multicolumn{5}{l}{\textbf{\boldmath EDM~\citep{karras2022elucidating}, CIFAR-10 ($32\times32$, cond) $\cdot$ Heun}} \\
14.0M & 35 & 3.26 & 3.08 & \textbf{\boldmath$-$0.18} \\
55.7M & 35 & 1.79 & 2.54 & +0.75 \\
\midrule
\addlinespace
\multicolumn{5}{l}{\textbf{\boldmath EDM~\citep{karras2022elucidating}, FFHQ ($64\times64$) $\cdot$ Heun}} \\
15.5M & 79 & 3.89 & 4.88 & +0.99 \\
61.8M & 79 & 2.39 & 4.17 & +1.78 \\
\midrule
\addlinespace
\multicolumn{5}{l}{\textbf{\boldmath EDM~\citep{karras2022elucidating}, AFHQv2 ($64\times64$) $\cdot$ Heun}} \\
15.5M & 79 & 3.27 & 3.50 & +0.23 \\
61.8M & 79 & 1.96 & 3.00 & +1.04 \\
\midrule
\addlinespace
\multicolumn{5}{l}{\textbf{\boldmath Rectified Flow~\citep{liu2023flow}, CIFAR-10 ($32\times32$) $\cdot$ RK45}} \\
15.5M & 302 / 332 & 5.40 & 4.03 & \textbf{\boldmath$-$1.37} \\
61.8M & 127 / 308 & 2.58 & 3.77 & +1.19 \\
\midrule
\addlinespace
\multicolumn{5}{l}{\textbf{\boldmath JiT~\citep{li2026back}, ImageNet ($256\times256$) $\cdot$ Heun}} \\
131.3M & 198 & 3.66 & 5.58 & +1.92 \\
\bottomrule
\end{tabular}

\end{table}

\begin{figure*}[t]
    \centering
    \includegraphics[width=\textwidth]{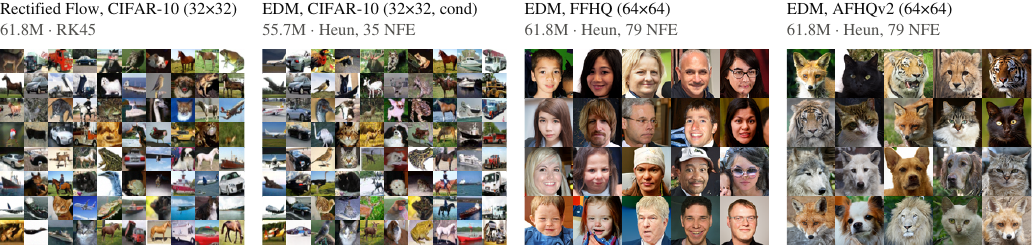}
    \caption{CIFAR-10, FFHQ and AFHQv2 uncurated samples from our \method.}
    \label{fig:samples}
\end{figure*}

\begin{figure}[t]
    \centering
    \includegraphics[width=\linewidth]{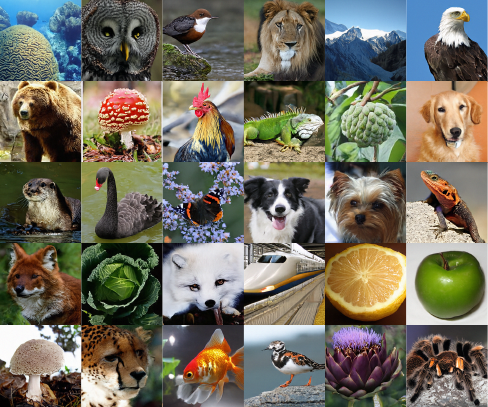}
    \caption{ImageNet $256\times256$ curated samples from our \method\ with JiT-B, Heun's solver at 198 NFE and guidance 4.0 for $t\in[0.1,1]$.}
    \label{fig:jit_curated}
\end{figure}

Tab.~\ref{tab:students_vs_teachers} compares \method\ with standard training of the same network, sampled with the same solver. For students 4$\times$ smaller than the teacher, we train these networks with the settings of \method\ (App.~\ref{appendix:competing}), and at the teacher's size, we report the published teachers and JiT-B/16~\citep{li2026back}. For JiT, the student uses a guidance scale of 4.0, which minimizes its FID (App.~\ref{appendix:cfg}), and standard training its published scale of 3.0. For the smaller students, \method\ outperforms standard training on CIFAR-10 by 0.38 FID, by 0.18 with class labels and by 1.37 with Rectified Flow, whereas standard training performs better on FFHQ and AFHQv2 at $64\times64$, by 0.99 and 0.23. At the teacher's size, \method\ students differ from standard training by 0.75 on class-conditional CIFAR-10 to 1.92 FID on ImageNet at $256\times 256$. There, \method\ stores 1.8 GB of weights instead of 252 GB of images (Tab.~\ref{tab:storage_main}). Fig.~\ref{fig:samples} shows uncurated samples of \method\ on CIFAR-10, FFHQ and AFHQv2, and Fig.~\ref{fig:jit_curated} curated samples on ImageNet.

\subsection{Comparison with training on generated images}\label{subsec:multi_step}
\begin{table*}[t]
  \centering
  \footnotesize
  \setlength{\tabcolsep}{4pt}
  \caption{\method\ vs. training on generated images in FID$\downarrow$. Column ``Data'' denotes stored images with ``\dataR '' for real, ``\dataG '' for generated and \dataN\ for none. \textbf{Bold} denotes best performance excluding the teacher.}
  \label{tab:stored_pairs}
  \begin{tabular}{cl rrc rrc rrc}
\toprule
 & & \multicolumn{3}{c}{\textbf{\boldmath EDM~\citep{karras2022elucidating}, $32\times32$ $\cdot$ Heun}} & \multicolumn{3}{c}{\textbf{\boldmath EDM~\citep{karras2022elucidating}, $64\times64$ $\cdot$ Heun}} & \multicolumn{3}{c}{\textbf{\boldmath Rectified Flow~\citep{liu2023flow}, $32\times32$ $\cdot$ RK45}} \\
\cmidrule(lr){3-5}\cmidrule(lr){6-8}\cmidrule(lr){9-11}
Data & Model & Params & NFE & CIFAR-10 & Params & NFE & FFHQ & Params & NFE & CIFAR-10 \\
\midrule
\dataR & Teacher & 55.7M & 35 & 1.97 & 61.8M & 79 & 2.39 & 61.8M & 127 & 2.58 \\
\dataG & SlimFlow (2-RF)~\citep{zhu2024slimflow} & 15.7M & 35 & 3.69 & 15.7M & 79 & 5.68 & 15.7M & -- & 4.51 \\
\cmidrule{1-11}
\dataN & \method\ (ours) & 14.0M & 35 & \textbf{3.56} & 15.5M & 79 & \textbf{4.88} & 15.5M & 332 & \textbf{4.03} \\
\bottomrule
\end{tabular}

\end{table*}

Tab.~\ref{tab:stored_pairs} compares \method\ with the multi-step 2-RF students of SlimFlow~\citep{zhu2024slimflow}, which, like \method, learn from a pretrained teacher from random initialization and without real data, but train on 50K to 200K stored pairs of noise and teacher-generated images. \method\ outperforms SlimFlow (2-RF) by 0.13 FID on CIFAR-10 and 0.80 FID on FFHQ with the EDM teachers, and by 0.48 FID on CIFAR-10 with the Rectified Flow teacher, while using no stored images.

\subsection{Comparison with step reduction methods}\label{subsec:step_reduction}
\begin{table}[t]
  \centering
  \footnotesize
  \setlength{\tabcolsep}{2.4pt}
  \caption{\method\ vs. step reduction methods trained from random initialization without stored images, in FID$\downarrow$. Column ``Data'' denotes stored images with ``\dataR '' for real and \dataN\ for none. \textbf{Bold} denotes best performance per section excluding the teacher. $^*$Heun's solver.}
  \label{tab:small_students}
  \begin{tabular}{clrrcc}
\toprule
Data & Model & Params & NFE & & \\
\midrule
\multicolumn{4}{l}{\textbf{\boldmath EDM~\citep{karras2022elucidating}, $32\times32$}} & \multicolumn{2}{c}{CIFAR-10} \\
\cmidrule(lr){5-6}
 & & & & uncond. & cond. \\
\cmidrule(lr){5-5}\cmidrule(lr){6-6}
\dataR & Teacher & 55.7M & 35$^*$ & 1.97 & 1.79 \\
\dataN & DMD2 (w/o GAN, rand)~\citep{yin2024improved} & 14.0M & 1 & 84.33 & 83.15 \\
\dataN & SiD (rand)~\citep{zhou2024score} & 14.0M & 1 & 351.25 & 234.14 \\
\cmidrule{1-6}
\dataN & \method\ (ours) & 14.0M & 35$^*$ & \textbf{3.56} & \textbf{3.08} \\
\midrule
\multicolumn{4}{l}{\textbf{\boldmath EDM~\citep{karras2022elucidating}, $64\times64$}} & FFHQ & AFHQv2 \\
\cmidrule(lr){5-5}\cmidrule(lr){6-6}
\dataR & Teacher & 61.8M & 79$^*$ & 2.39 & 1.96 \\
\dataN & DMD2 (w/o GAN, rand)~\citep{yin2024improved} & 15.5M & 1 & 99.09 & 156.35 \\
\dataN & SiD (rand)~\citep{zhou2024score} & 15.5M & 1 & 313.41 & 302.36 \\
\cmidrule{1-6}
\dataN & \method\ (ours) & 15.5M & 79$^*$ & \textbf{4.88} & \textbf{3.50} \\
\bottomrule
\end{tabular}

\end{table}

Tab.~\ref{tab:small_students} shows that step reduction methods do not extend to training a student without teacher-based initialization or stored images. We trained DMD2~\citep{yin2024improved}, without its GAN loss, and SiD~\citep{zhou2024score} from random initialization without stored images and they reach only 83.15 to 156.35 and 234.14 to 351.25 FID, respectively, whereas \method\ attains 3.08 to 4.88. Initialized from the teacher's weights, thus at the teacher's size, the same DMD2 configuration attains 6.9 to 13.7 FID, showing that these methods depend on teacher-based initialization, while local supervision along the teacher's trajectories succeeds where distribution-level objectives fail.

\subsection{Effectiveness on discrete-time teachers}\label{subsec:dkdm}
\begin{table*}[t]
  \centering
  \footnotesize
  \setlength{\tabcolsep}{4pt}
  \caption{\method\ on discrete-time teachers in FID$\downarrow$. Column ``Data'' denotes stored images with ``\dataR '' for real and \dataN\ for none. \textbf{Bold} denotes best performance excluding the teacher.}
  \label{tab:dkdm_benchmarks}
  \begin{tabular}{cl rrcc rrc rrcc}
\toprule
 & & \multicolumn{4}{c}{\textbf{\boldmath ADM~\citep{ning2023input}, $32\times32$ $\cdot$ Heun + churn}} & \multicolumn{3}{c}{\textbf{\boldmath ADM~\citep{ning2023input}, $64\times64$ $\cdot$ Heun + churn}} & \multicolumn{4}{c}{\textbf{\boldmath LDM~\citep{rombach2022high}, $256\times256$ $\cdot$ DDIM}} \\
\cmidrule(lr){3-6}\cmidrule(lr){7-9}\cmidrule(lr){10-13}
Data & Model & Params & NFE & CIFAR-10 & ImageNet & Params & NFE & CelebA & Params & NFE & CelebA-HQ & FFHQ \\
\midrule
\dataR & Teacher & 57.1M & 99 & 2.60 & 2.66 & 295.1M & 99 & 2.33 & 274.1M & 50 & 5.58 & 5.38 \\
\dataN & DKDM~\citep{xiang2025dkdm} & 14.3M & 99 & 6.59 & 8.55 & 57.2M & 99 & \textbf{6.74} & 89.5M & 50 & \textbf{8.56} & 11.03 \\
\cmidrule{1-13}
\dataN & \method\ (ours) & 14.3M & 99 & \textbf{4.16} & \textbf{7.28} & 57.2M & 99 & 7.99 & 89.5M & 50 & 8.61 & \textbf{10.80} \\
\bottomrule
\end{tabular}

\end{table*}

Although \method\ is formulated for continuous-time teachers, it applies, through the noise-level parameterization of EDM~\citep{karras2022elucidating}, to the discrete-time ADM and LDM teachers to which DKDM is restricted. Tab.~\ref{tab:dkdm_benchmarks} compares the two methods with the same teachers, student architectures and solvers, on CIFAR-10, ImageNet and CelebA with ADM and on CelebA-HQ and FFHQ with LDM. With these solvers, DKDM's released students improve on their published FID in every setting, by 0.13 to 2.97 (App.~\ref{appendix:competing}). \method\ outperforms DKDM by 2.43 FID on CIFAR-10, 1.27 on ImageNet and 0.23 on FFHQ, and is within 0.05 FID on CelebA-HQ. On CelebA, DKDM attains a 1.25 lower FID. \method\ attains these results at a disadvantage by construction, as it rounds every noise level to one of the teacher's discrete timesteps, which removes the continuum it is designed to supervise, and follows deterministic sampling trajectories, whereas DKDM natively follows the teacher's stochastic denoising chain.

\subsection{Effectiveness across architectures}\label{subsec:cross_arch}
\begin{table}[t]
  \centering
  \footnotesize
  \setlength{\tabcolsep}{4pt}
  \caption{\method\ cross-architecture performance on CIFAR-10.}
  \label{tab:cross_arch}
  \begin{tabular}{lccc}
\toprule
\multicolumn{2}{l}{Teacher} & \multicolumn{2}{c}{Student} \\
\cmidrule(r){1-2}\cmidrule(l){3-4}
Model & FID$\downarrow$ & \begin{tabular}[b]{@{}c@{}}\textbf{CNN} (14.0M)\\DDPM++~\citep{song2020score}\end{tabular} & \begin{tabular}[b]{@{}c@{}}\textbf{ViT} (19.3M)\\DiT~\citep{peebles2023scalable,xiang2025dkdm}\end{tabular} \\
\midrule
\textbf{CNN} (EDM~\citep{karras2022elucidating}, 55.7M) & 1.97 & 3.56 & 7.42 \\
\textbf{ViT} (U-ViT-S/2~\citep{bao2023all}, 44.3M) & 3.11 & 5.32 & 6.72 \\
\bottomrule
\end{tabular}

\end{table}

Tab.~\ref{tab:cross_arch} shows that \method\ effectively trains CNN and ViT students, evaluated with deterministic Heun sampling at 35 NFE, from CNN and ViT teachers on CIFAR-10. In this experiment, the architecture of the student matters more than its correspondence with the teacher's, as the CNN student outperforms the DiT student under both teachers, with 3.56 against 7.42 FID from EDM and 5.32 against 6.72 FID from U-ViT-S/2. For context, DKDM reports 17.11 FID for the same DiT student of a discrete-time ADM teacher~\citep{xiang2025dkdm}.

\subsection{Computational and storage costs}\label{subsec:cost}
\begin{table}[t]
  \centering
  \footnotesize
  \setlength{\tabcolsep}{4pt}
  \caption{Training throughput in images/s on one A100 GPU. ``Ratio'' denotes \method\ relative to standard training. Official implementation of standard training in FP32$^\dagger$ and FP16$^\ddagger$.}
  \label{tab:compute_main}
  \begin{tabular}{lrrrr}
\toprule
Teacher, dataset & Params & Standard training & \method & Ratio \\
\midrule
\multirow{4}{*}{\begin{tabular}[c]{@{}l@{}}EDM~\citep{karras2022elucidating},\\CIFAR-10\end{tabular}} & 14.0M & 1,839 & 914 & 0.50$\times$ \\
 & 55.7M & 860 & 589 & 0.68$\times$ \\
 & 55.7M & 234$^\dagger$ & 589 & 2.52$\times$ \\
 & 55.7M & 776$^\ddagger$ & 589 & 0.76$\times$ \\
\midrule
\multirow{2}{*}{\begin{tabular}[c]{@{}l@{}}Rectified Flow~\citep{liu2023flow},\\CIFAR-10\end{tabular}} & 61.8M & 1,136 & 824 & 0.73$\times$ \\
 & 61.8M & 607$^\dagger$ & 824 & 1.36$\times$ \\
\midrule
LDM~\citep{rombach2022high}, FFHQ & 89.5M & 109 & 153 & 1.40$\times$ \\
\midrule
JiT~\citep{li2026back}, ImageNet & 131.3M & 668 & 133 & 0.20$\times$ \\
\bottomrule
\end{tabular}

\end{table}

\begin{table}[t]
  \centering
  \footnotesize
  \setlength{\tabcolsep}{4pt}
  \caption{Disk storage of teacher weights and training images. ``Ratio'' denotes images relative to weights.}
  \label{tab:storage_main}
  \begin{tabular}{llrrr}
\toprule
 & & \multicolumn{2}{c}{Storage (GB)} & \\
\cmidrule(lr){3-4}
Teacher & Dataset & Weights & Images & Ratio \\
\midrule
EDM~\citep{karras2022elucidating} & CIFAR-10 ($32\times32$) & 0.22 & 0.15 & 0.7$\times$ \\
EDM~\citep{karras2022elucidating} & FFHQ ($64\times64$) & 0.25 & 0.86 & 3.5$\times$ \\
LDM~\citep{rombach2022high} & FFHQ ($256\times256$) & 1.10 & 11.8 & 11$\times$ \\
JiT-L/16~\citep{li2026back} & ImageNet ($256\times256$) & 1.84 & 251.9 & 137$\times$ \\
\bottomrule
\end{tabular}

\end{table}

Tab.~\ref{tab:compute_main} compares the training throughput of \method\ with that of standard training of the same student and batch size (further details in App.~\ref{appendix:cost}). Unrolling the teacher lowers it to 0.50 to 0.73$\times$ on EDM and Rectified Flow and to 0.20$\times$ on JiT, whose guided teacher is 3.5$\times$ larger than its student, and raises the peak memory by at most 5\%, whereas on LDM \method\ is 1.40$\times$ faster, as it encodes no training images. \method\ also trains 2.52$\times$ and 1.36$\times$ faster than the official implementations of EDM and Rectified Flow in single precision, and attains 0.76$\times$ the throughput of EDM in half precision. Tab.~\ref{tab:storage_main} shows that the weights of the teacher occupy more space than the 50K images of CIFAR-10, but 11$\times$ and 137$\times$ less than FFHQ and ImageNet at $256\times256$.

\subsection{Ablation study}\label{subsec:ablation}
\begin{table}[t]
  \centering
  \footnotesize
  \setlength{\tabcolsep}{4pt}
  \caption{Performance in FID$\downarrow$ by discretization scheme.}
  \label{tab:randomization}
  \begin{tabular}{lcccc}
\toprule
 & \multicolumn{2}{c}{EDM ($\mu_\Omega=60$)} & \multicolumn{2}{c}{Rectified Flow ($\mu_\Omega=30$)} \\
\cmidrule(lr){2-3}\cmidrule(lr){4-5}
Scheme & $M=b$ & $M=3b$ & $M=b$ & $M=3b$ \\
\midrule
Fixed grid & 99.53 & 4.91 & 12.66 & 15.66 \\
Random positions & 30.06 & 4.95 & 6.51 & \textbf{4.91} \\
DDM & \textbf{5.17} & \textbf{4.76} & \textbf{5.44} & 5.23 \\
\bottomrule
\end{tabular}

\end{table}

\begin{table}[t]
  \centering
  \footnotesize
  \caption{FID$\downarrow$ by unroll solver and Euler's throughput vs. Heun's.}
  \label{tab:unroll}
  \begin{tabular}{lllr}
\toprule
Teacher & Heun & Euler & Ratio \\
\midrule
EDM, $\{20,100\}$ & \textbf{4.76} & 5.32 & 1.33$\times$ \\
Rectified Flow, $\{10,50\}$ & \textbf{5.23} & 5.39 & 1.26$\times$ \\
\bottomrule
\end{tabular}

\end{table}

\textbf{Discretization scheme.} Tab.~\ref{tab:randomization} compares the three discretization schemes at equal mean step count. Without a pool ($M=b$), the fixed grid and random positions fail on EDM (99.53 and 30.06 FID) and degrade on Rectified Flow (12.66 and 6.51), with one noise level per batch and concentrated levels, respectively (Fig.~\ref{fig:coverage}b), whereas \method, with independent levels, attains 5.17 and 5.44. The other two schemes succeed only with a pool of $M=3b$, which renders their levels independent, whereas \method\ does not depend on it and improves only moderately, to 4.76 and 5.23. The fixed grid then suffices on EDM (4.91) but not on Rectified Flow (15.66), where RK45 evaluates the student between its 30 supervised levels, far fewer than the 774 of \method. Random positions are comparable on EDM (4.95), where their unbounded steps cost slightly more on curved trajectories, and better on Rectified Flow (4.91). \method\ is the only scheme that succeeds on both teachers with and without a pool.

\begin{figure}[t]
    \centering
    \includegraphics[width=\linewidth]{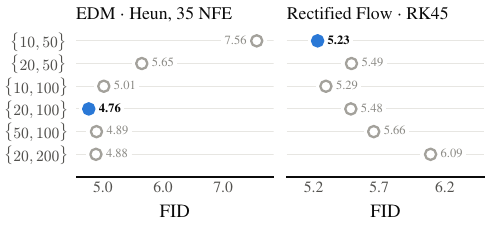}
    \caption{Step-count bounds of \method\ on CIFAR-10. FID of the 14.0M EDM and 15.5M Rectified Flow students against the bounds $\{n_0,n_1\}$, with a pool of $M=3b$. Filled denote adopted bounds.}
    \label{fig:ablation}
\end{figure}
\textbf{Pool size.} Although \method\ does not require a pool for independent noise levels, a pool of $M=3b$ still reduces its FID from 5.17 to 4.76 on EDM and from 5.44 to 5.23 on Rectified Flow, at a memory cost of $2b$ additional noisy samples, a small fraction of the training memory. A larger pool of $M=10b$ yields 4.66 on EDM but 5.53 on Rectified Flow, and is therefore not adopted.

\textbf{Unroll solver.} Unrolling the teacher with Euler's instead of Heun's solver (Tab.~\ref{tab:unroll}) halves the teacher evaluations, at throughput ratios of 1.33$\times$ and 1.26$\times$, but raises the FID from 4.76 to 5.32 on EDM and from 5.23 to 5.39 on Rectified Flow. This is consistent with the larger discretization error of first-order steps on the curved trajectories of EDM, which affects both the states and the targets of the student (App.~\ref{appendix:random_positions}).

\textbf{Step-count bounds.} In Fig.~\ref{fig:ablation}, the lower bound $n_0$ sets the largest step, $1/n_0$. Raising $n_0$ from 10 to 20 reduces the FID on EDM from 7.56 to 5.65 at $n_1=50$, as smaller steps improve the targets on curved trajectories, but raises it on Rectified Flow, whose straighter trajectories tolerate larger steps. Raising $n_1$ to 200 increases the density but reduces the number of trajectories by about 45\% and raises the FID on both. Our bounds, $\{20,100\}$ for EDM and $\{10,50\}$ for Rectified Flow, attain the lowest FID (4.76 and 5.23).

\section{Conclusions}
This work demonstrates that the weights of a pretrained continuous-time diffusion or flow model suffice to train a new denoising model from random initialization, without storing any training image. By unrolling the teacher along trajectories of random step counts, \method\ supervises the student at independent, approximately continuous and dense noise levels and imposes no constraint on the architecture or size of the student. Its students approach and, in some setups, even outperform standard training, and surpass both training on teacher-generated samples and step reduction methods under the same constraints. Overall, these results establish released teachers as a viable source of supervision when the training data are too large to store or unavailable.

\section*{Acknowledgements}
This work is partially funded by the Horizon Europe projects ELLIOT (GA No. 101214398) and lorAI (GA No. 101136646). We acknowledge the EuroHPC Joint Undertaking for awarding us access to Leonardo at CINECA, Italy. Google Cloud Platform resources were provided by the National Infrastructures for Research and Technology GRNET and funded by the EU Recovery and Resilience Facility.

{
    \small
    \bibliographystyle{ieeenat_fullname}
    \bibliography{main}
}

\clearpage
\setcounter{page}{1}
\maketitlesupplementary
\renewcommand{\topfraction}{0.9}\renewcommand{\dbltopfraction}{0.9}\renewcommand{\bottomfraction}{0.8}
\renewcommand{\textfraction}{0.07}\renewcommand{\floatpagefraction}{0.7}\renewcommand{\dblfloatpagefraction}{0.7}
\setcounter{topnumber}{3}\setcounter{dbltopnumber}{3}\setcounter{bottomnumber}{2}\setcounter{totalnumber}{5}

This supplement details the method, the experimental setup and further results.

\section{Teacher parameterization and unrolling}\label{appendix:teacher_step}\label{appendix:velocity}
\textbf{Velocity parameterization.} Our teachers differ in their prediction target and noise schedule. To accommodate them in a unified framework, we transform every model's output to velocity (cf. Sec.~\ref{subsec:background}). We express flow models on the linear path and diffusion models in EDM's formulation~\citep{karras2022elucidating} as
\begin{equation}\label{eq:coordinates}
\begin{array}{@{}l@{\qquad}l@{}}
    \mathbf{x}_t=t\,\mathbf{x}+(1-t)\,\mathbf{z}, & t:0\to1,\\[2pt]
    \mathbf{x}_\sigma=\mathbf{x}+\sigma\,\mathbf{z}, & \sigma:\sigma_{\max}\to0,
\end{array}
\end{equation}
where the arrows give the direction of sampling, from the prior $p_0$ to the data, with $p_0=\mathcal{N}(\mathbf{0},\mathbf{I})$ and $p_0=\mathcal{N}(\mathbf{0},\sigma_{\max}^2\mathbf{I})$, respectively. Using Eq.~\eqref{eq:velocity} we derive the teacher's velocity per family as
\begin{equation}\label{eq:teacher_velocity}
    v^T=
    \begin{cases}
        \mathbf{v}_\theta & \text{Rectified Flow},\\[3pt]
        \dfrac{\mathbf{x}_\theta-\mathbf{x}_t}{1-t} & \text{JiT},\\[8pt]
        \dfrac{\mathbf{x}_t-\mathbf{x}_\theta}{\sigma} & \text{EDM},\\[8pt]
        \mathbf{z}_\theta & \text{ADM, LDM, U-ViT}
    \end{cases}
\end{equation}
Rectified Flow~\citep{liu2023flow} predicts the velocity itself, while JiT~\citep{li2026back} and EDM~\citep{karras2022elucidating} predict the data, on the linear path and in EDM's formulation, respectively. The ADM teachers of DDPM-IP~\citep{ning2023input}, the LDM teachers~\citep{rombach2022high} and U-ViT~\citep{bao2023all} predict the noise on the variance-preserving path $(\mathbf{x}+\sigma\mathbf{z})/\sqrt{1+\sigma^2}$. The student's velocity, $v^S(\phi;\mathbf{x}_t,t)$, follows similarly from its own prediction, $\mathbf{x}_\phi$, $\mathbf{z}_\phi$ or $\mathbf{v}_\phi$.

\textbf{Schedule and step.} Algs.~\ref{alg:schedule} and~\ref{alg:teacher_step} detail the two teacher functions of Alg.~\ref{alg:teacher_sampling}, both applied element-wise over a batch. Flow models use the time $t\in[0,1]$ of the linear path, and diffusion models the noise level $\sigma$ of EDM's formulation, spaced by EDM's schedule~\citep{karras2022elucidating} with $\rho=7$ and $(\sigma_{\min},\sigma_{\max})=(0.002,80)$ for EDM, $(0.006425,80)$ for the ADM teachers of DDPM-IP and U-ViT, and $(0.038759,80)$ for the LDM teachers. ADM, LDM and U-ViT receive the rescaled state $\mathbf{x}/\sqrt{1+\sigma^2}$ at the time at which their own noise schedule reaches $\sigma$, rounded to the nearest of the 1,000 discrete timesteps of ADM and LDM~\citep{karras2022elucidating}. For conditional teachers, $v^T$ takes the condition $c$, and for unconditional ones $c$ is omitted. When the teacher is unrolled with classifier-free guidance~\citep{ho2022classifier}, as the JiT teacher, its state advances with the guided velocity $v^T_w(\theta;\mathbf{x},t,c)=v^T(\theta;\mathbf{x},t,\bar{c})+w\,\big(v^T(\theta;\mathbf{x},t,c)-v^T(\theta;\mathbf{x},t,\bar{c})\big)$, where $\bar{c}$ is the null condition and $w$ the guidance scale inside the guidance interval, with $w=1$ outside it, whereas the target remains the conditional velocity $v^T$. Without guidance, $w=1$ and $v^T_w=v^T$.

\begin{algorithm}[h]
\caption{\textsc{Schedule}$(\xi,\omega)$, the noise level at step $\xi$ of a trajectory of $\omega$ steps.}
\label{alg:schedule}
\begin{algorithmic}[1]
\If{the teacher is a flow model}
  \State \Return $\xi/\omega$
\ElsIf{the teacher is a diffusion model}
  \If{$\xi=\omega$}
    \State \Return $0$
  \EndIf
  \State \Return $\big(\sigma_{\max}^{1/\rho}+\frac{\xi}{\omega-1}(\sigma_{\min}^{1/\rho}-\sigma_{\max}^{1/\rho})\big)^{\rho}$
\EndIf
\end{algorithmic}
\end{algorithm}

\begin{algorithm}[h]
\caption{\textsc{TeacherStep}$(v^T;\mathbf{x},t,t',c)$, one Euler or Heun step of the teacher from $t$ to $t'$, returning the new state and the target velocity. The state advances with the guided velocity $v^T_w$, which equals $v^T$ without guidance.}
\label{alg:teacher_step}
\begin{algorithmic}[1]
\State $\mathbf{v}_1\gets v^T(\theta;\mathbf{x},t,c),\ \mathbf{v}^w_1\gets v^T_w(\theta;\mathbf{x},t,c)$
\State $\mathbf{x}'\gets\mathbf{x}+(t'-t)\,\mathbf{v}^w_1$
\If{the solver is Euler, or $t'$ ends the trajectory}
  \State \Return $(\mathbf{x}',\mathbf{v}_1)$
\EndIf
\State $\mathbf{v}_2\gets v^T(\theta;\mathbf{x}',t',c),\ \mathbf{v}^w_2\gets v^T_w(\theta;\mathbf{x}',t',c)$
\State \Return $\big(\mathbf{x}+(t'-t)\,\frac{\mathbf{v}^w_1+\mathbf{v}^w_2}{2},\ \frac{\mathbf{v}_1+\mathbf{v}_2}{2}\big)$
\end{algorithmic}
\end{algorithm}

\section{Noise levels of the training batches}\label{appendix:batch}\label{appendix:levels}
This section supports the statements of Sec.~\ref{subsec:ddm} on the noise levels drawn from the noisy sample pool.

\textbf{Setting.} Every trajectory draws its number of steps $\Omega$ independently, with probability $p_\omega=P(\Omega=\omega)$ of $\omega$ steps and mean $\mu_\Omega=\sum_\omega p_\omega\,\omega$. For \method, $\Omega\sim\mathcal{U}\{n_0,n_1\}$, with $p_\omega=1/(n_1-n_0+1)$ for every $\omega\in\{n_0,\dots,n_1\}$ and $\mu_\Omega=(n_0+n_1)/2$, the mean of the integers from $n_0$ to $n_1$. For a fixed grid of $n$ steps, $p_n=1$ and $\mu_\Omega=n$. Let $M=b$, and call each of the $b$ entries of the batch a batch position. At every iteration, the teacher advances the state of every batch position by one step. Over the course of training, a batch position thus holds the consecutive states of a trajectory, at the positions $\xi=0,1,\dots,\omega-1$, then the states of a new trajectory that starts from noise, and so on. We write $\tau=\xi/\omega$ for the normalized position of the state held by a batch position. With a pool of $M>b$ samples, the same holds for every sample of the pool, which advances whenever it is drawn into the batch.

\textbf{Proposition \ref*{appendix:levels}.1.} \textit{Over the course of training, the proportion of iterations at which a batch position holds a state with $\tau\le t$ converges to $P(\tau\le t)$ of Eq.~\eqref{eq:position_cdf}, which satisfies $t<P(\tau\le t)\le t+1/\mu_\Omega$ for all $t\in[0,1)$.}

\textit{Proof.} Consider one batch position and the sequence of trajectories that it holds, the $j$-th of which has $\Omega_j$ steps. The $j$-th trajectory occupies the batch position for $\Omega_j$ iterations, one for each of its positions. For $t\in[0,1)$, its positions with $\xi/\Omega_j\le t$ are $\xi=0,1,\dots,\lfloor\Omega_j t\rfloor$, and hence
\begin{equation}
    R_j=\lfloor\Omega_j t\rfloor+1
\end{equation}
of its $\Omega_j$ iterations have $\tau\le t$. The starts of the trajectories at the batch position form a renewal process~\citep[Definition 3.1.1, p. 98]{ross1996stochastic}, whose $j$-th interval lasts $\Omega_j$ iterations and earns the reward $R_j$. The pairs $(\Omega_j,R_j)$ are independent and identically distributed, since $R_j$ depends only on $\Omega_j$. By the renewal-reward theorem~\citep[Theorem 3.6.1, p. 133]{ross1996stochastic}, the proportion of iterations with $\tau\le t$ converges to $E[R]/E[\Omega]=\frac{1}{\mu_\Omega}\sum_\omega(\lfloor\omega t\rfloor+1)\,p_\omega$, which is Eq.~\eqref{eq:position_cdf}. For the bounds, every $\omega$ satisfies $\omega t<\lfloor\omega t\rfloor+1\le\omega t+1$. Multiplying by $p_\omega/\mu_\Omega$ and summing over $\omega$ gives
\begin{equation}
    \frac{t}{\mu_\Omega}\sum_\omega p_\omega\,\omega<P(\tau\le t)\le\frac{t}{\mu_\Omega}\sum_\omega p_\omega\,\omega+\frac{1}{\mu_\Omega}\sum_\omega p_\omega,
\end{equation}
which is $t<P(\tau\le t)\le t+1/\mu_\Omega$, since $\sum_\omega p_\omega\,\omega=\mu_\Omega$ and $\sum_\omega p_\omega=1$. $\square$

For a fixed grid of $n$ steps, Eq.~\eqref{eq:position_cdf} gives $P(\tau\le t)=(\lfloor nt\rfloor+1)/n$. The bound of Proposition~\ref*{appendix:levels}.1 therefore holds for both a fixed grid and \method. The two differ in the set of values that $\tau$ takes, the $n$ positions $\{0,1/n,\dots,(n-1)/n\}$ of a fixed grid against the $c_t$ positions of Eq.~\eqref{eq:levels} for \method, rather than in the distance of their distributions from the uniform one. Proposition~\ref*{appendix:levels}.1 describes one batch position over the course of training, however, and not the states that the $b$ batch positions hold at one iteration.

\textbf{Proposition \ref*{appendix:levels}.2.} \textit{Let $\Omega\sim\mathcal{U}\{n_0,n_1\}$ with $n_1>n_0$, and let all trajectories start at iteration $k=0$. For batch position $i$ at iteration $k$, let $\xi^{(i)}_k$ be the step of the state it holds, $\Omega^{(i)}_k$ the number of steps of its trajectory and $\tau^{(i)}_k=\xi^{(i)}_k/\Omega^{(i)}_k$. At every iteration $k$, the positions $\tau^{(1)}_k,\dots,\tau^{(b)}_k$ are independent and identically distributed, and their common distribution $P(\tau_k\le t)=P(\tau^{(i)}_k\le t)$ converges to $P(\tau\le t)$ of Eq.~\eqref{eq:position_cdf} as $k\to\infty$.}

\textit{Proof.} Each batch position draws its own numbers of steps, independently of the other batch positions and from the same distribution. The positions $\tau^{(1)}_k,\dots,\tau^{(b)}_k$ are therefore independent and identically distributed at every iteration. For the limit, it therefore suffices to consider one batch position, and we drop the index $i$. Let $s_k$ denote the probability that a trajectory starts at this batch position at iteration $k$, with $s_0=1$. For $k\ge n_1$, the batch position holds step $\xi$ of a trajectory of $\omega>\xi$ steps at iteration $k$ exactly when a trajectory starts at iteration $k-\xi$ and draws $\omega$ steps. The number of steps is drawn independently of the start, which gives
\begin{equation}
    P(\xi_k=\xi,\ \Omega_k=\omega)=s_{k-\xi}\,p_\omega.
\end{equation}
The event $\tau_k\le t$ consists of the pairs $(\xi,\omega)$ with $\xi\le\omega t$, that is, of the steps $\xi=0,1,\dots,\lfloor\omega t\rfloor$ for every $\omega$, and therefore
\begin{equation}
\begin{aligned}
    P(\tau_k\le t)&=\sum_\omega\sum_{\xi=0}^{\lfloor\omega t\rfloor}P(\xi_k=\xi,\ \Omega_k=\omega)\\
    &=\sum_\omega p_\omega\sum_{\xi=0}^{\lfloor\omega t\rfloor}s_{k-\xi}.
\end{aligned}
\end{equation}
A batch position starts one trajectory every $\mu_\Omega$ iterations on average. Blackwell's theorem~\citep[Theorem 3.4.1, p. 110]{ross1996stochastic} states that, after enough iterations, $s_k$ settles to $1/\mu_\Omega$, provided that the possible numbers of steps have no common divisor larger than 1. The numbers of steps of \method\ include $n_0$ and $n_0+1$, which have no common divisor, and hence $s_{k-\xi}\to1/\mu_\Omega$ for every $\xi$. Each inner sum has $\lfloor\omega t\rfloor+1$ terms, which yields
\begin{equation}
    P(\tau_k\le t)\to\frac{1}{\mu_\Omega}\sum_\omega\big(\lfloor\omega t\rfloor+1\big)\,p_\omega,
\end{equation}
which is Eq.~\eqref{eq:position_cdf}. $\square$

Within a batch, the positions are thus asymptotically distributed as the positions that one batch position holds over the course of training. For a fixed grid of $n$ steps, in contrast, all trajectories start at the iterations $0,n,2n,\dots$, and every batch position holds the position $(k\bmod n)/n$ at iteration $k$. Then $s_k$ is 1 at the multiples of $n$ and 0 otherwise, and the batch keeps a single noise level at every iteration, although Proposition~\ref*{appendix:levels}.1 holds over the course of training.

\textbf{Corollary \ref*{appendix:levels}.1.} \textit{For \method\ with $\Omega\sim\mathcal{U}\{n_0,n_1\}$, $n_1>n_0$ and $M=b$, the positions within a batch become independent samples of a distribution that differs from $\mathcal{U}(0,1)$ by at most $1/\mu_\Omega=2/(n_0+n_1)$ at every $t$.}

\textit{Proof.} By Proposition~\ref*{appendix:levels}.2, the positions within a batch become independent samples of the distribution of Eq.~\eqref{eq:position_cdf}, and by Proposition~\ref*{appendix:levels}.1, $|P(\tau\le t)-t|\le1/\mu_\Omega$, with $\mu_\Omega=(n_0+n_1)/2$. $\square$

The bound is $1/60$ for $\{n_0,n_1\}=\{20,100\}$ and tends to 0 as $n_1$ grows.

\begin{figure*}[!tp]
    \centering
    \includegraphics[width=\textwidth]{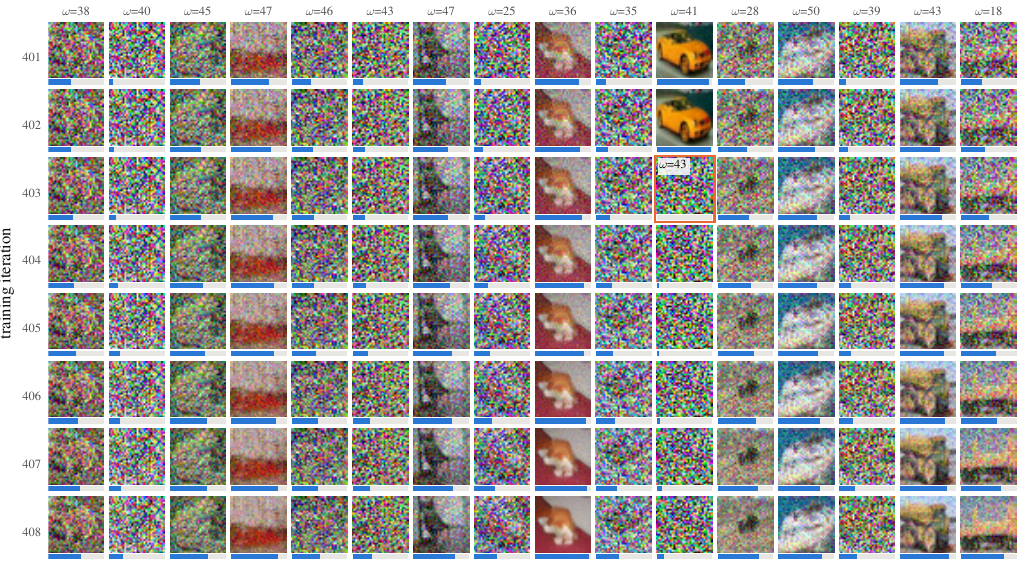}
    \caption{Formation of a training batch by \method, for the Rectified Flow CIFAR-10 teacher with $\Omega\sim\mathcal{U}\{10,50\}$ and a pool of $M=b=16$ noisy samples, after 400 warmup iterations. Columns are the samples of the batch and rows eight consecutive training iterations. At every iteration, each sample advances by one teacher step along its own trajectory, and the bar beneath it shows its noise level $t$. An orange square marks a sample that completed its trajectory and restarted from noise. Labels give the step count $\omega$ of each trajectory.}
    \label{fig:batch_trajectory}
\end{figure*}

\textbf{Illustration.} Fig.~\ref{fig:batch_trajectory} illustrates Proposition~\ref*{appendix:levels}.2 with a batch of $b=16$ noisy samples over eight consecutive training iterations, generated by the Rectified Flow CIFAR-10 teacher with $\Omega\sim\mathcal{U}\{10,50\}$ (Sec.~\ref{sec:setup}) and a pool of only $M=b$ samples. At every iteration, each sample advances by one teacher step along its own trajectory, and a sample that completes its trajectory restarts from noise. As the trajectories have different step counts, the samples complete them at different iterations. After a warmup, the batch therefore spans the noise range at every iteration, whereas all samples of a fixed grid without a pool share a single noise level (Fig.~\ref{fig:coverage}b).

\textbf{Number of distinct levels.} Grids of different step counts share some positions, such as $1/2=10/20=25/50$, and Eq.~\eqref{eq:levels} counts each position once. Most positions are nevertheless distinct, and the union of the grids contains $c_t=774$ levels for $\{10,50\}$ and 3,044 for $\{20,100\}$, against at most 50 and 100 for a single grid. For the EDM and U-ViT teachers, whose noise levels lie at $\xi/(\omega-1)$ (Alg.~\ref{alg:schedule}), the union contains 3,005 levels for $\{20,100\}$. The discrete-time ADM and LDM teachers are defined on 1,000 discrete timesteps, to the nearest of which every noise level is rounded (App.~\ref{appendix:teacher_step}). The union therefore cannot exceed the 992 and 995 steps up to the one nearest to $\sigma_{\max}=80$. For $\{20,100\}$, \method\ reaches 848 and 906 of them which narrows its advantage in temporal coverage on these teachers.

\section{Architectures and training settings}\label{appenix:architectures}\label{appendix:training}
Tab.~\ref{tab:arch_config} lists the teachers and students of all experiments, with the resolution and configuration of every network. The 14.0M and 15.5M students halve the channel width of their teacher, and the 19.8M and 25.6M students reduce it to 96 with fewer blocks per resolution. The latent students operate on the $3\times64\times64$ latents of the teacher's VQ-f4 autoencoder and share its frozen decoder, whose 33.0M parameters are not counted.

\begin{table*}[!tp]
    \centering
    \footnotesize
    \setlength{\tabcolsep}{5pt}
    \caption{Teachers and students of all experiments and their network configurations.}
    \label{tab:arch_config}
    \begin{tabular}{lrcccccccccc}
\toprule
 & & \multicolumn{4}{c}{U-Net} & \multicolumn{5}{c}{Transformer} & \\
\cmidrule(lr){3-6}\cmidrule(lr){7-11}
Model & Params & \begin{tabular}[b]{@{}c@{}}Base\\ch.\end{tabular} & \begin{tabular}[b]{@{}c@{}}Ch.\\mult.\end{tabular} & \begin{tabular}[b]{@{}c@{}}Blocks/\\res.\end{tabular} & \begin{tabular}[b]{@{}c@{}}Attn.\\res.\end{tabular} & Patch & Depth & Hidden & Heads & MLP & Dropout \\
\midrule
\multicolumn{12}{l}{\textbf{Continuous-time teachers}} \\
\begin{tabular}[t]{@{}l@{}}EDM, CIFAR-10\\$32\times32$, uncond.\ and class-cond.\end{tabular} & 55.7M & 128 & 2,2,2 & 4 & 16 & -- & -- & -- & -- & -- & -- \\
\quad DDPM++ student & 14.0M & 64 & 2,2,2 & 4 & 16 & -- & -- & -- & -- & -- & 0.1 \\
\quad DDPM++ student & 19.8M & 96 & 2,2,2 & 2 & 16 & -- & -- & -- & -- & -- & 0.1 \\
\quad DDPM++ student & 25.6M & 96 & 2,2,2 & 3 & 16 & -- & -- & -- & -- & -- & 0.1 \\
\quad DDPM++ student & 55.7M & 128 & 2,2,2 & 4 & 16 & -- & -- & -- & -- & -- & 0.1 \\
\begin{tabular}[t]{@{}l@{}}EDM, FFHQ / AFHQv2\\$64\times64$\end{tabular} & 61.8M & 128 & 1,2,2,2 & 4 & 16 & -- & -- & -- & -- & -- & -- \\
\quad DDPM++ student & 15.5M & 64 & 1,2,2,2 & 4 & 16 & -- & -- & -- & -- & -- & 0.1 \\
\quad DDPM++ student & 61.8M & 128 & 1,2,2,2 & 4 & 16 & -- & -- & -- & -- & -- & 0.1 \\
\begin{tabular}[t]{@{}l@{}}Rectified Flow, CIFAR-10\\$32\times32$\end{tabular} & 61.8M & 128 & 1,2,2,2 & 4 & 16 & -- & -- & -- & -- & -- & -- \\
\quad DDPM++ student & 15.5M & 64 & 1,2,2,2 & 4 & 16 & -- & -- & -- & -- & -- & 0.15 \\
\quad DDPM++ student & 61.8M & 128 & 1,2,2,2 & 4 & 16 & -- & -- & -- & -- & -- & 0.15 \\
\begin{tabular}[t]{@{}l@{}}JiT-L/16, ImageNet\\$256\times256$, class-cond.\end{tabular} & 459.1M & -- & -- & -- & -- & 16 & 24 & 1024 & 16 & 4 & -- \\
\quad JiT-B/16 student & 131.3M & -- & -- & -- & -- & 16 & 12 & 768 & 12 & 4 & 0 \\
\midrule
\multicolumn{12}{l}{\textbf{ADM and LDM teachers}} \\
\begin{tabular}[t]{@{}l@{}}ADM, CIFAR-10 / ImageNet\\$32\times32$\end{tabular} & 57.1M & 128 & 1,2,2,2 & 3 & 16,8 & -- & -- & -- & -- & -- & -- \\
\quad ADM student & 14.3M & 64 & 1,2,2,2 & 3 & 16,8 & -- & -- & -- & -- & -- & 0.1 \\
\begin{tabular}[t]{@{}l@{}}ADM, CelebA\\$64\times64$\end{tabular} & 295.1M & 192 & 1,2,3,4 & 3 & 32,16,8 & -- & -- & -- & -- & -- & -- \\
\quad ADM student & 57.2M & 96 & 1,2,3,4 & 2 & 32,16,8 & -- & -- & -- & -- & -- & 0.1 \\
\begin{tabular}[t]{@{}l@{}}LDM-VQ-4 UNet, CelebA-HQ / FFHQ\\$256\times256$, latent $3\times64\times64$\end{tabular} & 274.1M & 224 & 1,2,3,4 & 2 & 32,16,8 & -- & -- & -- & -- & -- & -- \\
\quad UNet student & 89.5M & 128 & 1,2,3,4 & 2 & 32,16,8 & -- & -- & -- & -- & -- & 0 \\
\midrule
\multicolumn{12}{l}{\textbf{Cross-architecture pairs}} \\
\begin{tabular}[t]{@{}l@{}}EDM, CIFAR-10\\$32\times32$\end{tabular} & 55.7M & 128 & 2,2,2 & 4 & 16 & -- & -- & -- & -- & -- & -- \\
\quad DiT student & 19.3M & -- & -- & -- & -- & 2 & 7 & 384 & 6 & 4 & 0 \\
\begin{tabular}[t]{@{}l@{}}U-ViT-S/2, CIFAR-10\\$32\times32$\end{tabular} & 44.3M & -- & -- & -- & -- & 2 & 13 & 512 & 8 & 4 & -- \\
\quad DDPM++ student & 14.0M & 64 & 2,2,2 & 4 & 16 & -- & -- & -- & -- & -- & 0.1 \\
\quad DiT student & 19.3M & -- & -- & -- & -- & 2 & 7 & 384 & 6 & 4 & 0 \\
\bottomrule
\end{tabular}

\end{table*}

Tab.~\ref{tab:training} lists the training settings of all runs. Every teacher advances the pool with Heun's solver, except in the comparison of solvers of Sec.~\ref{subsec:ablation}. The students are trained with Adam~\citep{kingma2015adam} without weight decay, with $(\beta_1,\beta_2)=(0.9,0.999)$, except $\beta_2=0.95$ for JiT, after a linear warmup of the learning rate, and their weights are averaged by an EMA whose decay is $\min\{d,(1+i)/(10+i)\}$ at update $i$ for the final decay $d$ of the table, except for JiT, which uses $d$ throughout. Teacher and student run in bfloat16 mixed precision, and the loss is computed in single precision. The guided students receive the null label with probability 0.1, and the JiT teacher is unrolled with classifier-free guidance~\citep{ho2022classifier} at scale 3.0 for $t\in[0.1,1]$. The DiT student computes attention in single precision, since in bfloat16 it diverges at the peak learning rate.
\definecolor{ablation}{HTML}{008300}
\begin{table*}[!tp]
    \centering
    \footnotesize
    \setlength{\tabcolsep}{2.2pt}
    \caption{Training settings of all runs. \textcolor{ablation}{Runs of the ablation study (Sec.~\ref*{subsec:ablation}).}}
    \label{tab:training}
    \begin{tabular}{llllrrrrrlll}
        \toprule
        Dataset & Teacher & Student & $\{n_0,n_1\}$ & $K$ & $b$ & $M$ & LR & Warmup & EMA & Dropout & GPUs \\
        \midrule
        CIFAR-10 & EDM & 14.0M, 19.8M, 25.6M, 55.7M & $\{20,100\}$ & 400K & 1024 & $3b$ & 2e-4 & 5K & 0.9999 & 0.10 & 4 \\
        CIFAR-10 (cond.) & EDM & 14.0M, 55.7M & $\{20,100\}$ & 400K & 1024 & $3b$ & 2e-4 & 5K & 0.9999 & 0.10 & 4 \\
        FFHQ ($64\times64$) & EDM & 15.5M, 61.8M & $\{20,100\}$ & 400K & 1024 & $3b$ & 2e-4 & 5K & 0.9999 & 0.10 & 4, 8 \\
        AFHQv2 ($64\times64$) & EDM & 15.5M, 61.8M & $\{20,100\}$ & 400K & 1024 & $3b$ & 2e-4 & 5K & 0.9999 & 0.10 & 4, 8 \\
        CIFAR-10 & Rectified Flow & 15.5M, 61.8M & $\{10,50\}$ & 400K & 1024 & $3b$ & 2e-4 & 5K & 0.9999 & 0.15 & 4 \\
        ImageNet ($256\times256$, cond.) & JiT-L/16 & JiT-B/16 131.3M & $\{10,50\}$ & 500K & 2048 & $4b$ & 3.5e-4 & 10K & 0.9997 & 0.00 & 8 \\
        \midrule
        CIFAR-10 & ADM & 14.3M & $\{20,100\}$ & 400K & 1024 & $3b$ & 2e-4 & 5K & 0.9999 & 0.10 & 4 \\
        ImageNet ($32\times32$) & ADM & 14.3M & $\{20,100\}$ & 500K & 2048 & $3b$ & 2e-4 & 5K & 0.9999 & 0.10 & 4 \\
        CelebA ($64\times64$) & ADM & 57.2M & $\{20,100\}$ & 400K & 512 & $3b$ & 2e-4 & 5K & 0.9999 & 0.10 & 4 \\
        CelebA-HQ ($256\times256$) & LDM-VQ-4 & 89.5M & $\{20,100\}$ & 500K & 512 & $3b$ & 2e-4 & 5K & 0.9999 & 0.00 & 4 \\
        FFHQ ($256\times256$) & LDM-VQ-4 & 89.5M & $\{20,100\}$ & 500K & 512 & $3b$ & 2e-4 & 5K & 0.9999 & 0.00 & 4 \\
        \midrule
        CIFAR-10 & EDM & DiT 19.3M & $\{20,100\}$ & 400K & 1024 & $3b$ & 2e-4 & 5K & 0.9999 & 0.00 & 4 \\
        CIFAR-10 & U-ViT-S/2 & DDPM++ 14.0M & $\{20,100\}$ & 400K & 1024 & $3b$ & 2e-4 & 5K & 0.9999 & 0.10 & 4 \\
        CIFAR-10 & U-ViT-S/2 & DiT 19.3M & $\{20,100\}$ & 400K & 1024 & $3b$ & 2e-4 & 5K & 0.9999 & 0.00 & 4 \\
        \midrule
        \textcolor{ablation}{CIFAR-10} & \textcolor{ablation}{EDM} & \textcolor{ablation}{14.0M} & \textcolor{ablation}{varied} & \textcolor{ablation}{80K} & \textcolor{ablation}{1024} & \textcolor{ablation}{$b$, $3b$, $10b$} & \textcolor{ablation}{2e-4} & \textcolor{ablation}{5K} & \textcolor{ablation}{0.9999} & \textcolor{ablation}{0.10} & \textcolor{ablation}{4} \\
        \textcolor{ablation}{CIFAR-10} & \textcolor{ablation}{Rectified Flow} & \textcolor{ablation}{15.5M} & \textcolor{ablation}{varied} & \textcolor{ablation}{100K} & \textcolor{ablation}{1024} & \textcolor{ablation}{$b$, $3b$, $10b$} & \textcolor{ablation}{2e-4} & \textcolor{ablation}{5K} & \textcolor{ablation}{0.9999} & \textcolor{ablation}{0.15} & \textcolor{ablation}{4} \\
        \bottomrule
    \end{tabular}
\end{table*}

\section{FID calculation}\label{appendix:fid_stats}
We compute the FID between 50,000 generated samples and the reference statistics of Tab.~\ref{tab:fid_stats}, without augmentation, using torch-fidelity (\url{https://github.com/LTH14/torch-fidelity}), whose Inception-v3 is a PyTorch translation of the same TensorFlow model that EDM~\citep{karras2022elucidating} uses. Following EDM, we evaluate the final checkpoint of every \method\ student and of our standard training three times with different sampling seeds and report the lowest FID. The one exception is the FFHQ student of the LDM teacher, whose evaluation at 480K iterations attains a slightly lower FID than at 500K and is reported instead. The ablations report the evaluation during training at their final iteration. Tab.~\ref{tab:fid_stats} lists the reference statistics of every evaluation. Where the teacher's authors release statistics of its training set, we use them. For the ADM and LDM teachers, neither DKDM nor the authors of the teachers release statistics, so we rebuild them from the training sets exactly as DKDM prepares them for its evaluation, image for image, and extract the same Inception features as for the other datasets. For CelebA, these are all 202,599 aligned images, and for CelebA-HQ and FFHQ the 25,000 and 60,000 images of the training split of LDM~\citep{rombach2022high}. The released statistics come from the repositories of EDM (\url{https://github.com/NVlabs/edm}), Rectified Flow (\url{https://github.com/gnobitab/RectifiedFlow}) and JiT (\url{https://github.com/LTH14/JiT}), and the rebuilt ones follow the preparation code of DKDM (\url{https://github.com/iLearn-Lab/CVPR25-DKDM}).
\begin{table}[H]
    \centering
    \footnotesize
    \setlength{\tabcolsep}{3pt}
    \caption{FID reference statistics of all evaluations.}
    \label{tab:fid_stats}
    \begin{tabular}{@{}lll@{}}
        \toprule
        Dataset & Teachers & Source \\
        \midrule
        CIFAR-10 & \begin{tabular}[t]{@{}l@{}}EDM, ADM,\\U-ViT\end{tabular} & EDM repository \\
        CIFAR-10 & Rectified Flow & Rectified Flow repository \\
        FFHQ, AFHQv2 ($64\times64$) & EDM & EDM repository \\
        ImageNet ($256\times256$) & JiT & JiT repository \\
        \midrule
        ImageNet ($32\times32$) & ADM & Rebuilt, 1{,}281{,}167 images \\
        CelebA ($64\times64$) & ADM & Rebuilt, 202{,}599 images \\
        CelebA-HQ ($256\times256$) & LDM & Rebuilt, 25{,}000 images \\
        FFHQ ($256\times256$) & LDM & Rebuilt, 60{,}000 images \\
        \bottomrule
    \end{tabular}
\end{table}

\section{Competing methods}\label{appendix:competing}
\textbf{Standard training at smaller sizes.} For the smaller students of Tab.~\ref{tab:students_vs_teachers}, which have no published counterpart, we train the same networks with standard training on the training images of their teachers, namely, the 50K images of CIFAR-10 and, at $64\times64$, the 70K of FFHQ and the 15,803 of AFHQv2. The EDM students use the objective of EDM~\citep{karras2022elucidating}, with log-normally distributed noise levels and its loss weighting, and the Rectified Flow student that of Rectified Flow~\citep{liu2023flow}, with $t\sim\mathcal{U}(10^{-3},1)$ on the linear path and its velocity target. The network, batch size, number of iterations, optimizer, learning-rate schedule, EMA, numerical precision and evaluation are those of the corresponding \method\ student (App.~\ref{appendix:training}), hence only the training signal differs. We augment the images by horizontal flips only, as the non-leaky augmentation of EDM requires an additional network input for the augmentation parameters. EDM additionally tunes the learning rate, batch size, dropout rate and augmentation probability of each dataset, hence these models represent standard training under the settings of \method\ rather than its best attainable result. On CIFAR-10, their FID stops improving after about 240K iterations, whereas that of \method\ decreases until the end of training.

\textbf{Training of DMD2 and SiD.} We train DMD2~\citep{yin2024improved} and SiD~\citep{zhou2024score} with their official implementations. Both use the teacher as the real score and initialize the fake score from the teacher, and we train their generators from random initialization with the architecture of the \method\ students, DDPM++ with 64 base channels (14.0M at $32\times32$ and 15.5M at $64\times64$). DMD2 is trained with its distribution-matching loss and without its GAN loss, which requires real images. Both of its networks use AdamW with a learning rate of $10^{-5}$ and a weight decay of 0.01, with five fake-score updates per generator update, a generator input noise level of $\sigma=80$, gradient clipping at 10, 500 warmup iterations and a batch of 512 at $32\times32$ and 256 at $64\times64$, for 400K iterations. SiD, which requires no data by design, is trained with the official hyperparameters of each dataset, a batch of 256 with $\alpha=1.2$ on CIFAR-10 and of 512 with $\alpha=1.2$ on FFHQ and $\alpha=1.0$ on AFHQv2. We stop SiD after 100K iterations, as its generator collapses to a single image, or one per class, within its first few hundred iterations, remains collapsed in three of the four settings, and its FID shows no consistent improvement across evaluations every 20K iterations. The teacher-initialized DMD2 runs reported in Sec.~\ref{subsec:step_reduction} use the same configuration with the generator initialized from the teacher. Both methods are evaluated at their final iteration with the protocol of App.~\ref{appendix:fid_stats}.

\textbf{Evaluation of competing methods.} We report published numbers wherever they exist for the compared setting, namely standard training of EDM, Rectified Flow and JiT-B/16 at the teacher's size (Tab.~\ref{tab:students_vs_teachers}), the teachers of Tabs.~\ref{tab:stored_pairs}, \ref{tab:small_students} and~\ref{tab:cross_arch}, the student of \citet{dockhorn2023distilling} (Tab.~\ref{tab:dockhorn}), the 2-RF model of SlimFlow with the Rectified Flow teacher, and the teachers and students of DKDM as published by DKDM (Tab.~\ref{tab:dkdm_published}). Where no published number exists for the compared setting, we evaluate the released checkpoints with the protocol of App.~\ref{appendix:fid_stats}. This applies to the 2-RF models of SlimFlow with the EDM teachers, whose FIDs SlimFlow does not report, to the teachers and the released students of DKDM with the solvers of Tab.~\ref{tab:dkdm_benchmarks}, and to the teachers of Tab.~\ref{tab:students_vs_teachers_heun256} at 511 NFE, against the reference statistics of Tab.~\ref{tab:fid_stats}.

\begin{table*}[t]
  \centering
  \footnotesize
  \setlength{\tabcolsep}{4pt}
  \caption{FID$\downarrow$ of the released students of DKDM and of their teachers, as published by DKDM~\citep{xiang2025dkdm} and as measured by us with the solvers of Tab.~\ref{tab:dkdm_benchmarks}. ``$\Delta$'' denotes our measurement minus the published FID.}
  \label{tab:dkdm_published}
  \begin{tabular}{llr lrc lrc r}
\toprule
 & & & \multicolumn{3}{c}{Published} & \multicolumn{3}{c}{Solvers of Tab.~\ref{tab:dkdm_benchmarks}} & \\
\cmidrule(lr){4-6}\cmidrule(lr){7-9}
Teacher & Dataset & Params & Solver & NFE & FID & Solver & NFE & FID & $\Delta$ \\
\midrule
\multicolumn{10}{l}{\textbf{Students of DKDM}} \\
ADM~\citep{ning2023input} & CIFAR-10 ($32\times32$) & 14.3M & IDDPM & 50 & 9.56 & Heun + churn & 99 & 6.59 & $-$2.97 \\
ADM~\citep{ning2023input} & ImageNet ($32\times32$) & 14.3M & IDDPM & 50 & 11.33 & Heun + churn & 99 & 8.55 & $-$2.78 \\
ADM~\citep{ning2023input} & CelebA ($64\times64$) & 57.2M & IDDPM & 50 & 7.07 & Heun + churn & 99 & 6.74 & $-$0.33 \\
LDM~\citep{rombach2022high} & CelebA-HQ ($256\times256$) & 89.5M & DDIM & 200 & 8.69 & DDIM & 50 & 8.56 & $-$0.13 \\
LDM~\citep{rombach2022high} & FFHQ ($256\times256$) & 89.5M & DDIM & 200 & 11.53 & DDIM & 50 & 11.03 & $-$0.50 \\
\midrule
\multicolumn{10}{l}{\textbf{Teachers}} \\
ADM~\citep{ning2023input} & CIFAR-10 ($32\times32$) & 57.1M & IDDPM & 50 & 4.45 & Heun + churn & 99 & 2.60 & $-$1.85 \\
ADM~\citep{ning2023input} & ImageNet ($32\times32$) & 57.1M & IDDPM & 50 & 4.67 & Heun + churn & 99 & 2.66 & $-$2.01 \\
ADM~\citep{ning2023input} & CelebA ($64\times64$) & 295.1M & IDDPM & 50 & 4.43 & Heun + churn & 99 & 2.33 & $-$2.10 \\
LDM~\citep{rombach2022high} & CelebA-HQ ($256\times256$) & 274.1M & DDIM & 200 & 5.69 & DDIM & 50 & 5.58 & $-$0.11 \\
LDM~\citep{rombach2022high} & FFHQ ($256\times256$) & 274.1M & DDIM & 200 & 5.93 & DDIM & 50 & 5.38 & $-$0.55 \\
\bottomrule
\end{tabular}

\end{table*}

\section{Training progress}\label{appendix:convergence}
\begin{figure*}[!tp]
    \centering
    \includegraphics[width=\textwidth]{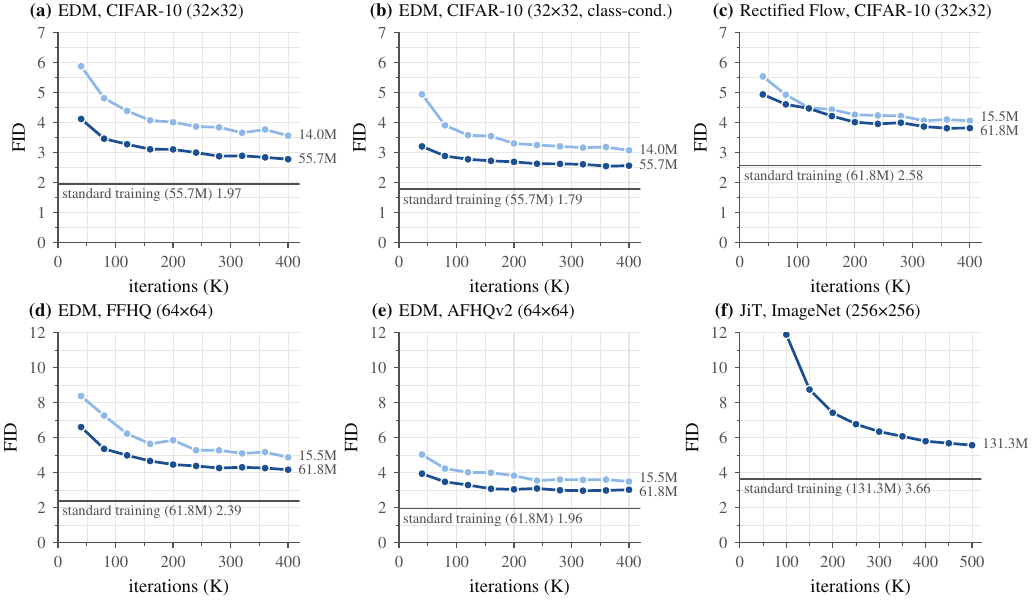}
    \caption{Training progress of \method\ students, FID (50K samples) against training iterations, one panel per teacher. Students are labelled by parameter count, and grey lines mark standard training at the FIDs their papers report, the teacher itself, and for JiT the JiT-B of the student's size trained on real data instead.}
    \label{fig:convergence}
\end{figure*}

Fig.~\ref{fig:convergence} shows the FID of \method\ students against training iterations for the teachers of Tab.~\ref{tab:students_vs_teachers}, together with the FIDs reported for standard training. Ten of the eleven students still improve at the end of training, by 2\% to 6\% of their FID per 100K iterations, and only the 61.8M AFHQv2 student has saturated, which suggests room for improvement with longer training.

\section{Guidance scale and comparison with standard training at 511 NFE}\label{appendix:cfg}\label{appendix:nfe}
\textbf{Guidance scale.} Tab.~\ref{tab:cfg_sweep} reports the FID and Inception Score~\citep{salimans2016improved} of the JiT-B/16 student across guidance scales. Its FID is lowest at 4.0, the scale of Tab.~\ref{tab:students_vs_teachers}, whereas its Inception Score increases with the scale.

\begin{table}[H]
    \centering
    \footnotesize
    \setlength{\tabcolsep}{4pt}
    \caption{FID and Inception Score (IS) of the \method\ JiT-B/16 student (131.3M) of the JiT-L/16 teacher on ImageNet at $256\times256$ against the guidance scale, with Heun's solver at 198 NFE and guidance for $t\in[0.1,1]$.}
    \label{tab:cfg_sweep}
    \begin{tabular}{lrrrrrr}
\toprule
Guidance scale & 3.0 & 3.5 & 4.0 & 4.5 & 5.0 & 5.5 \\
\midrule
FID$\downarrow$ & 7.05 & 5.94 & \textbf{5.58} & 5.62 & 5.87 & 6.19 \\
IS$\uparrow$ & 191.0 & 221.8 & 246.4 & 266.1 & 282.2 & 295.0 \\
\bottomrule
\end{tabular}

\end{table}

\textbf{Comparison with standard training at 511 NFE.} Tab.~\ref{tab:students_vs_teachers_heun256} compares the EDM students of Tab.~\ref{tab:students_vs_teachers} with their teachers at a higher sampling cost, Heun's solver with 256 steps (511 NFE). The difference to standard training decreases from 0.82 to 0.70 FID on CIFAR-10 and from 0.75 to 0.62 FID on class-conditional CIFAR-10, and changes by at most 0.01 FID on FFHQ and AFHQv2.

\begin{table}[H]
  \centering
  \footnotesize
  \setlength{\tabcolsep}{4pt}
  \caption{\method\ students at the size of their EDM teachers against standard training, both with Heun's solver at 256 steps (511 NFE). Column ``Data'' denotes stored images with ``\dataR '' for real and \dataN\ for none. Column ``$\Delta$'' denotes the FID difference with respect to standard training.}
  \label{tab:students_vs_teachers_heun256}
  \begin{tabular}{cllrll}
\toprule
Data & Model & Params & NFE & FID$\downarrow$ & $\Delta$ \\
\midrule
\addlinespace
\multicolumn{6}{l}{\textbf{\boldmath EDM, CIFAR-10 ($32\times32$)}} \\
\dataR & Standard training~\citep{karras2022elucidating} & 55.7M & 511 & 2.06 & -- \\
\dataN & \method & 55.7M & 511 & 2.76 & +0.70 \\
\midrule
\addlinespace
\multicolumn{6}{l}{\textbf{\boldmath EDM, CIFAR-10 ($32\times32$, cond)}} \\
\dataR & Standard training~\citep{karras2022elucidating} & 55.7M & 511 & 1.89 & -- \\
\dataN & \method & 55.7M & 511 & 2.51 & +0.62 \\
\midrule
\addlinespace
\multicolumn{6}{l}{\textbf{\boldmath EDM, FFHQ ($64\times64$)}} \\
\dataR & Standard training~\citep{karras2022elucidating} & 61.8M & 511 & 2.38 & -- \\
\dataN & \method & 61.8M & 511 & 4.17 & +1.79 \\
\midrule
\addlinespace
\multicolumn{6}{l}{\textbf{\boldmath EDM, AFHQv2 ($64\times64$)}} \\
\dataR & Standard training~\citep{karras2022elucidating} & 61.8M & 511 & 2.02 & -- \\
\dataN & \method & 61.8M & 511 & 3.05 & +1.03 \\
\bottomrule
\end{tabular}

\end{table}

\section{Computational and storage costs}\label{appendix:cost}
Tab.~\ref{tab:compute_cost} compares the training throughput and the peak GPU memory of \method\ with those of standard training of the same student at the same batch size. Both share the training step of the student, which compiles its blocks, runs its forward pass in bfloat16 mixed precision with the loss in single precision, and performs the optimizer step and the EMA update as fused operations over all parameters, and both are timed after warm-up iterations that absorb the compilation. They differ only in how the inputs and targets of the student are obtained. Standard training loads and decodes images at the training resolution at every iteration and, in latent space, encodes them with the teacher's autoencoder, whereas \method\ advances the pool with the teacher, which is compiled as a whole and evaluated in bfloat16 without gradients. In latent space, \method\ trains 1.40$\times$ faster than standard training on CelebA-HQ and FFHQ at $256\times256$, as it operates on the $3\times64\times64$ latents throughout, whereas standard training processes every image at full resolution. Pre-encoding the images would remove this cost, but the latents of the whole training set would then have to be stored, which training without stored images excludes. In pixel space, the teacher's evaluations dominate the cost of \method, which attains 0.42 to 0.73$\times$ the throughput of standard training, and 0.20$\times$ with the JiT student, whose teacher is 3.5$\times$ larger and is evaluated four times per iteration by Heun's solver with classifier-free guidance. Since the teacher runs without gradients, the memory overhead of \method\ is close to the teacher's weights, from 0.17 to 1.92GB, or 2 to 5\% of the memory of standard training in every setting.

Tab.~\ref{tab:original_cost} compares \method\ with the official training code of EDM and Rectified Flow, measured with the same protocol, GPU and batch size, each running its own training step, including its loss, preconditioning and data augmentation. Our implementation differs from both in three aspects. First, it runs the forward passes in bfloat16 mixed precision, with TF32 for the remaining single-precision matrix products, whereas the official implementations train in single precision, which EDM computes without TF32 and, as an option, in half precision. Second, it compiles the student's blocks, whereas both official implementations run without compilation. Third, it fuses the optimizer step and the EMA update over all parameters, whereas both official implementations update the EMA weights parameter by parameter, as EDM also does to remove non-finite gradient values. As a result, our implementation of standard training is 3.68$\times$ faster than the default configuration of EDM, 1.11$\times$ faster than its half-precision option and 1.87$\times$ faster than Rectified Flow. \method\ trains 2.52$\times$ faster than the default configuration of EDM and 1.36$\times$ faster than Rectified Flow, attains 0.76$\times$ the throughput of EDM in half precision, and requires 0.42 to 0.80$\times$ the peak memory of the official implementations. We omit JiT, whose official implementation already shares the main optimizations of ours. It compiles the same transformer blocks, runs the forward pass in bfloat16 mixed precision and trains with distributed data parallelism, and differs from ours only in an unfused optimizer step and an EMA update performed parameter by parameter.

\begin{table*}[!tp]
    \centering
    \footnotesize
    \setlength{\tabcolsep}{3pt}
    \caption{Training throughput and peak memory of one training step on one A100 (64GB) GPU, for \method\ and for standard training of the same student at the same batch size. Ratios denote \method\ relative to standard training.}
    \label{tab:compute_cost}\label{tab:memory_cost}
    \begin{tabular}{llllrrrrrrr}
\toprule
 & & & & & \multicolumn{3}{c}{Images/s} & \multicolumn{3}{c}{Peak memory (GB)} \\
\cmidrule(lr){6-8}\cmidrule(lr){9-11}
Teacher & Student & Dataset & Resolution & Batch & \multicolumn{1}{c}{\begin{tabular}[b]{@{}c@{}}Standard\\training\end{tabular}} & DDM & Ratio & \multicolumn{1}{c}{\begin{tabular}[b]{@{}c@{}}Standard\\training\end{tabular}} & DDM & Ratio \\
\midrule
EDM (55.7M) & DDPM++ (14.0M) & CIFAR-10 & $32\times32$ & 256 & 1,839 & 914 & 0.50$\times$ & 6.35 & 6.57 & 1.03$\times$ \\
EDM (55.7M) & DDPM++ (19.8M) & CIFAR-10 & $32\times32$ & 256 & 1,734 & 889 & 0.51$\times$ & 6.26 & 6.48 & 1.04$\times$ \\
EDM (55.7M) & DDPM++ (25.6M) & CIFAR-10 & $32\times32$ & 256 & 1,340 & 772 & 0.58$\times$ & 7.93 & 8.15 & 1.03$\times$ \\
EDM (55.7M) & DDPM++ (55.7M) & CIFAR-10 & $32\times32$ & 256 & 860 & 589 & 0.68$\times$ & 13.00 & 13.22 & 1.02$\times$ \\
EDM (55.7M) & DiT (19.3M) & CIFAR-10 & $32\times32$ & 256 & 2,656 & 1,111 & 0.42$\times$ & 7.98 & 8.19 & 1.03$\times$ \\
EDM (55.7M) & DDPM++ (14.0M) & CIFAR-10 (cond) & $32\times32$ & 256 & 1,834 & 918 & 0.50$\times$ & 6.35 & 6.57 & 1.03$\times$ \\
EDM (55.7M) & DDPM++ (55.7M) & CIFAR-10 (cond) & $32\times32$ & 256 & 872 & 592 & 0.68$\times$ & 12.99 & 13.21 & 1.02$\times$ \\
EDM (61.8M) & DDPM++ (15.5M) & FFHQ, AFHQv2 & $64\times64$ & 256 & 830 & 411 & 0.50$\times$ & 15.20 & 15.45 & 1.02$\times$ \\
EDM (61.8M) & DDPM++ (61.8M) & FFHQ, AFHQv2 & $64\times64$ & 128 & 369 & 253 & 0.69$\times$ & 15.91 & 16.16 & 1.02$\times$ \\
\midrule
Rectified Flow (61.8M) & DDPM++ (15.5M) & CIFAR-10 & $32\times32$ & 256 & 2,069 & 1,208 & 0.58$\times$ & 5.23 & 5.47 & 1.05$\times$ \\
Rectified Flow (61.8M) & DDPM++ (61.8M) & CIFAR-10 & $32\times32$ & 256 & 1,136 & 824 & 0.73$\times$ & 10.51 & 10.74 & 1.02$\times$ \\
\midrule
U-ViT-S/2 (44.3M) & DDPM++ (14.0M) & CIFAR-10 & $32\times32$ & 256 & 1,835 & 1,055 & 0.57$\times$ & 6.35 & 6.52 & 1.03$\times$ \\
U-ViT-S/2 (44.3M) & DiT (19.3M) & CIFAR-10 & $32\times32$ & 256 & 2,653 & 1,322 & 0.50$\times$ & 7.98 & 8.15 & 1.02$\times$ \\
\midrule
ADM (57.1M) & ADM (14.3M) & CIFAR-10 & $32\times32$ & 256 & 2,379 & 1,310 & 0.55$\times$ & 5.79 & 6.02 & 1.04$\times$ \\
ADM (57.1M) & ADM (14.3M) & ImageNet & $32\times32$ & 512 & 2,871 & 1,495 & 0.52$\times$ & 11.32 & 11.55 & 1.02$\times$ \\
ADM (295.1M) & ADM (57.2M) & CelebA & $64\times64$ & 128 & 352 & 174 & 0.49$\times$ & 26.93 & 28.05 & 1.04$\times$ \\
LDM-VQ-4 (274.1M) & UNet (89.5M) & CelebA-HQ, FFHQ & \begin{tabular}[t]{@{}l@{}}$256\times256$\\(latent $3\times64\times64$)\end{tabular} & 128 & 109 & 153 & 1.40$\times$ & 35.36 & 36.47 & 1.03$\times$ \\
\midrule
JiT-L (459.1M) & JiT-B (131.3M) & ImageNet & $256\times256$ & 256 & 668 & 133 & 0.20$\times$ & 45.83 & 47.75 & 1.04$\times$ \\
\bottomrule
\end{tabular}

\end{table*}

\begin{table*}[!tp]
    \centering
    \footnotesize
    \setlength{\tabcolsep}{3pt}
    \caption{Training throughput and peak memory of one training step on one A100 (64GB) GPU, for \method\ and for the official implementation of standard training of the same student at the same batch size. Column ``Precision'' denotes that of the official implementation, whereas \method\ trains with bfloat16 mixed precision. Ratios denote \method\ relative to the official implementation.}
    \label{tab:original_cost}
    \begin{tabular}{llllrlrrrrrr}
\toprule
 & & & & & & \multicolumn{3}{c}{Images/s} & \multicolumn{3}{c}{Peak memory (GB)} \\
\cmidrule(lr){7-9}\cmidrule(lr){10-12}
Teacher & Student & Dataset & Resolution & Batch & Precision & \multicolumn{1}{c}{\begin{tabular}[b]{@{}c@{}}Official\\standard training\end{tabular}} & DDM & Ratio & \multicolumn{1}{c}{\begin{tabular}[b]{@{}c@{}}Official\\standard training\end{tabular}} & DDM & Ratio \\
\midrule
EDM (55.7M) & DDPM++ (55.7M) & CIFAR-10 & $32\times32$ & 256 & FP32 & 234 & 589 & 2.52$\times$ & 31.42 & 13.22 & 0.42$\times$ \\
EDM (55.7M) & DDPM++ (55.7M) & CIFAR-10 & $32\times32$ & 256 & FP16 & 776 & 589 & 0.76$\times$ & 16.49 & 13.22 & 0.80$\times$ \\
\midrule
Rectified Flow (61.8M) & DDPM++ (61.8M) & CIFAR-10 & $32\times32$ & 256 & FP32 & 607 & 824 & 1.36$\times$ & 22.37 & 10.74 & 0.48$\times$ \\
\bottomrule
\end{tabular}

\end{table*}

Tab.~\ref{tab:storage} compares the disk storage of the teacher's weights, which \method\ requires, with that of the training images, which standard training requires. Weights are counted at 4 bytes per parameter and images as uncompressed 8-bit RGB at the training resolution. The ratio grows with the dataset, from below 1 on CIFAR-10 and AFHQv2 to 137 on ImageNet at $256\times256$. Compressed formats reduce the size of the images, as the JPEG archive of the ImageNet training set of about 138 GB, which still exceeds the teacher's weights 75 times, whereas released checkpoints, which also hold EMA weights and optimizer states, exceed the weights alone.

\begin{table*}[!tp]
    \centering
    \footnotesize
    \setlength{\tabcolsep}{3pt}
    \caption{Disk storage of the teacher's weights and of its training images. Ratio denotes the size of the images relative to that of the weights.}
    \label{tab:storage}
    \begin{tabular}{llrrrrr}
\toprule
 & & \multicolumn{2}{c}{Weights} & \multicolumn{2}{c}{Training images} & \\
\cmidrule(lr){3-4}\cmidrule(lr){5-6}
Teacher & Dataset & Params & Size (GB) & Number & Size (GB) & Ratio \\
\midrule
EDM & CIFAR-10, $32\times32$ & 55.7M & 0.22 & 50,000 & 0.15 & 0.7 \\
EDM, class-cond. & CIFAR-10, $32\times32$ & 55.7M & 0.22 & 50,000 & 0.15 & 0.7 \\
Rectified Flow & CIFAR-10, $32\times32$ & 61.8M & 0.25 & 50,000 & 0.15 & 0.6 \\
U-ViT-S/2 & CIFAR-10, $32\times32$ & 44.3M & 0.18 & 50,000 & 0.15 & 0.9 \\
ADM & CIFAR-10, $32\times32$ & 57.1M & 0.23 & 50,000 & 0.15 & 0.7 \\
EDM & AFHQv2, $64\times64$ & 61.8M & 0.25 & 15,803 & 0.19 & 0.8 \\
EDM & FFHQ, $64\times64$ & 61.8M & 0.25 & 70,000 & 0.86 & 3.5 \\
ADM & CelebA, $64\times64$ & 295.1M & 1.18 & 202,599 & 2.49 & 2.1 \\
ADM & ImageNet, $32\times32$ & 57.1M & 0.23 & 1,281,167 & 3.94 & 17 \\
LDM & CelebA-HQ, $256\times256$ & 274.1M & 1.10 & 25,000 & 4.92 & 4.5 \\
LDM & FFHQ, $256\times256$ & 274.1M & 1.10 & 60,000 & 11.8 & 11 \\
JiT-L/16 & ImageNet, $256\times256$ & 459.1M & 1.84 & 1,281,167 & 251.9 & 137 \\
\bottomrule
\end{tabular}

\end{table*}

\section{Samples generated by \method}\label{appenix:samples}
\begin{figure*}[!tp]
    \centering
    \includegraphics[width=\textwidth]{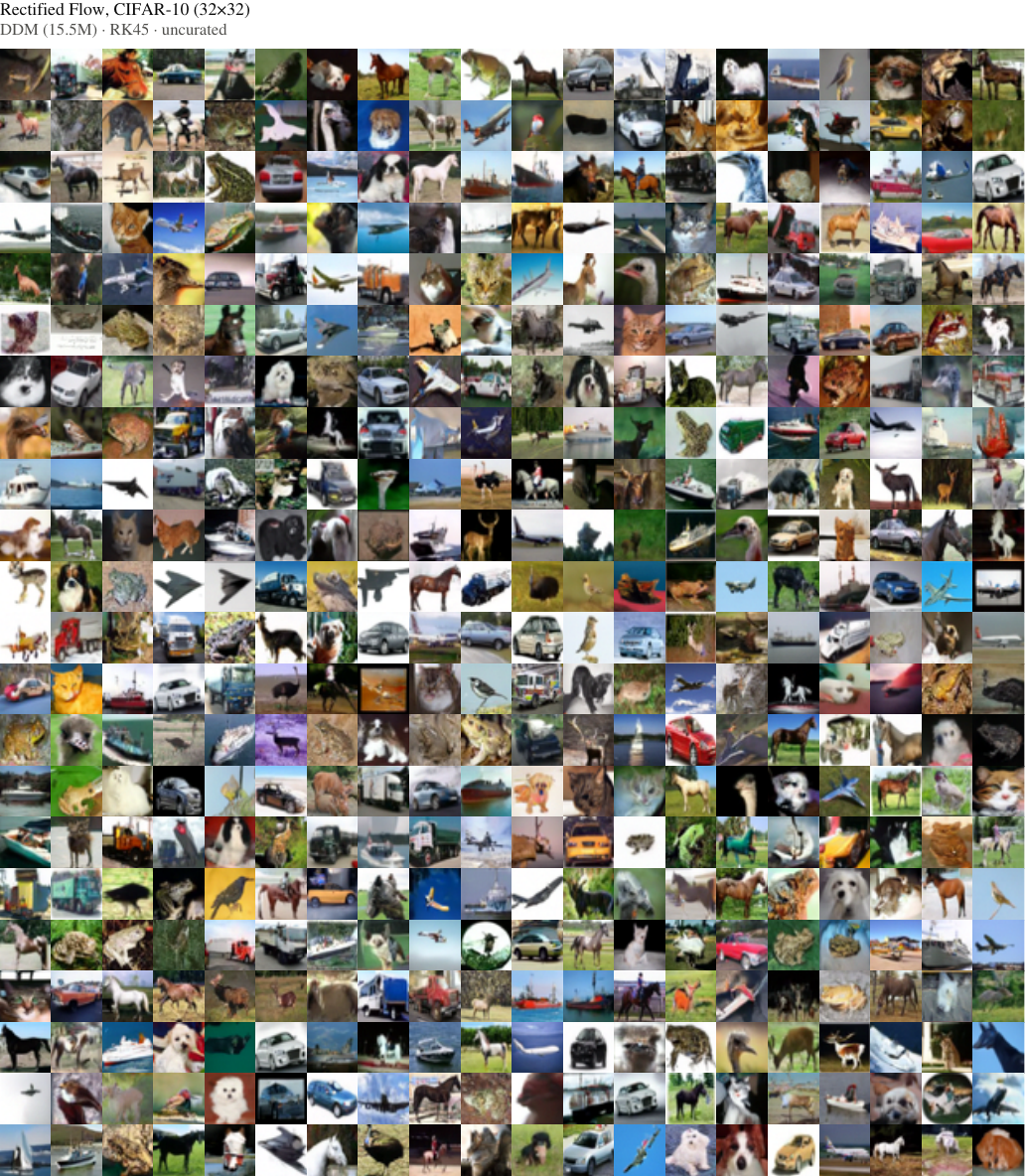}
    \caption{Uncurated samples of the 15.5M \method\ student of the Rectified Flow teacher on CIFAR-10, with RK45.}
    \label{fig:samples_small}
\end{figure*}

\begin{figure*}[!tp]
    \centering
    \includegraphics[width=\textwidth]{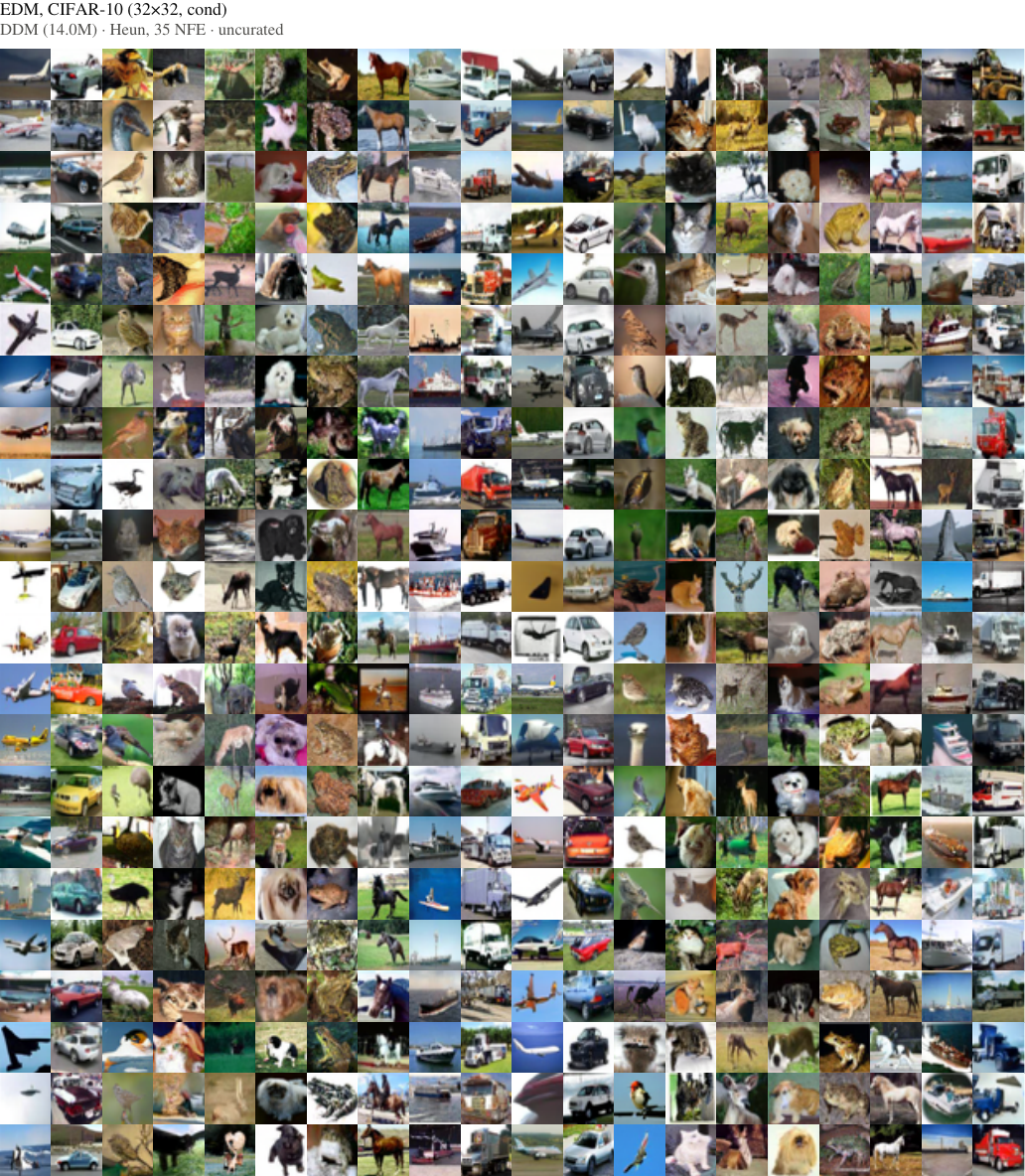}
    \caption{Uncurated samples of the 14.0M \method\ student of the class-conditional EDM teacher on CIFAR-10, with Heun's solver at 35 NFE. Each column shows one class.}
    \label{fig:samples_cond}
\end{figure*}

\begin{figure*}[!tp]
    \centering
    \includegraphics[width=\textwidth]{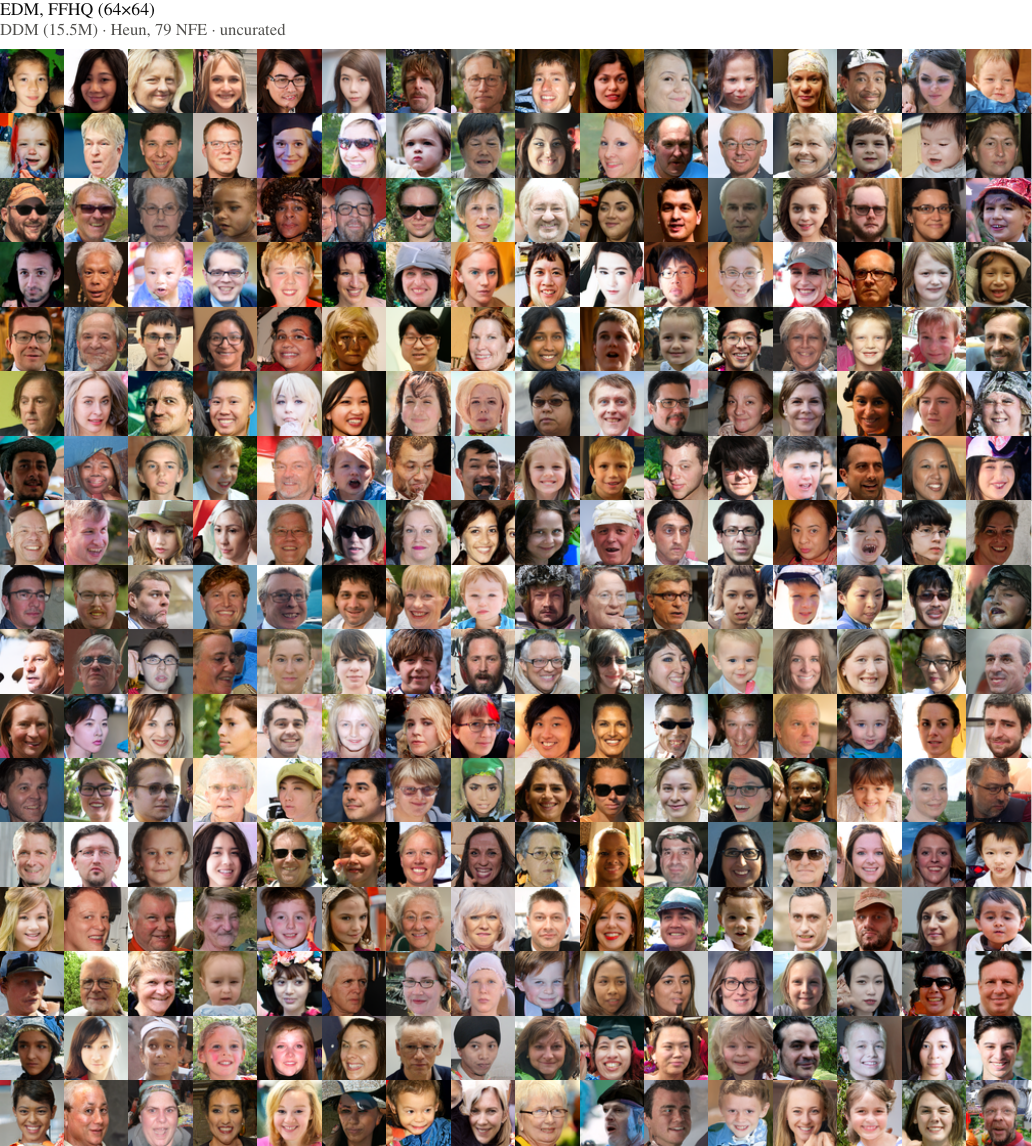}
    \caption{Uncurated samples of the 15.5M \method\ student of the EDM teacher on FFHQ at $64\times64$, with Heun's solver at 79 NFE.}
    \label{fig:samples_ffhq}
\end{figure*}

\begin{figure*}[!tp]
    \centering
    \includegraphics[width=\textwidth]{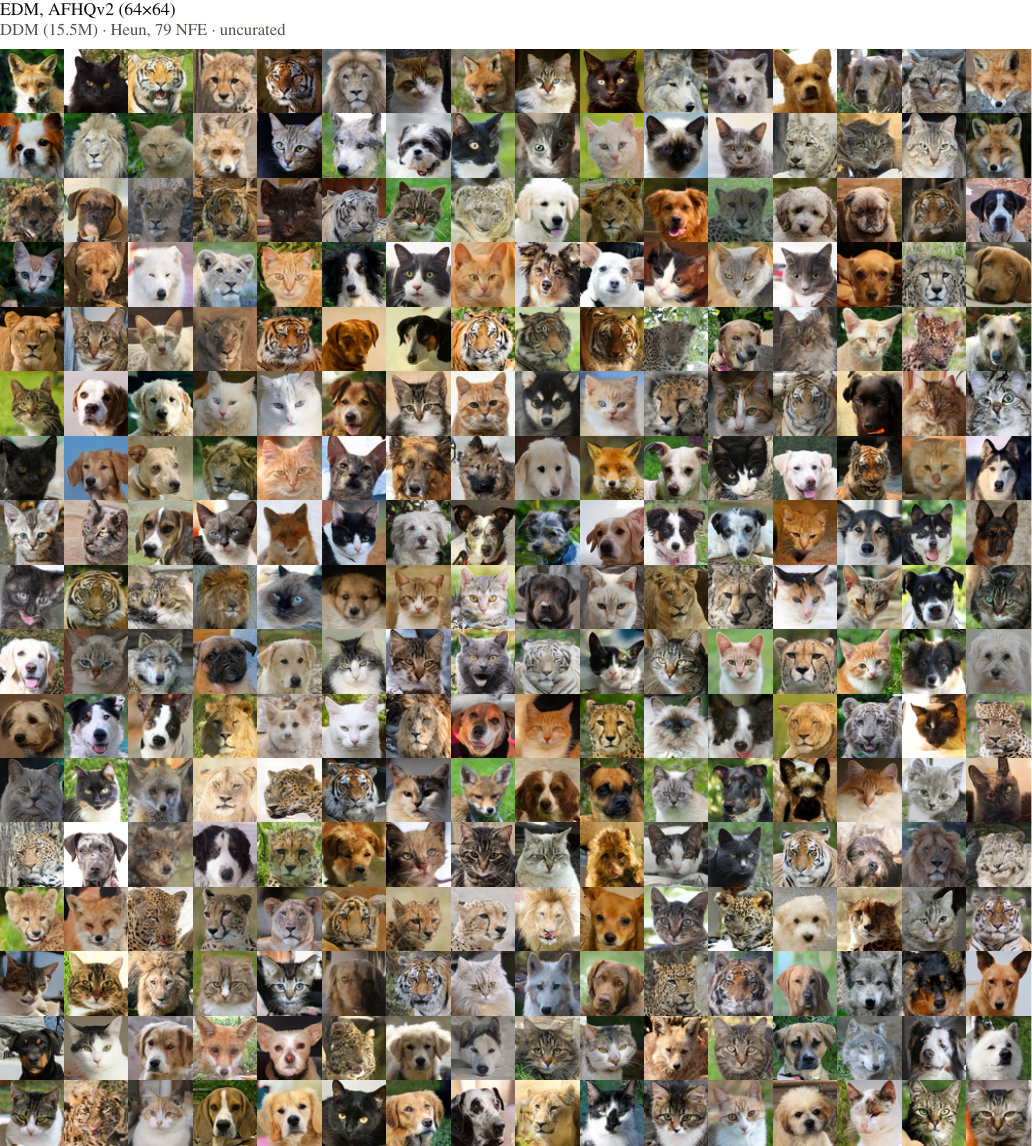}
    \caption{Uncurated samples of the 15.5M \method\ student of the EDM teacher on AFHQv2 at $64\times64$, with Heun's solver at 79 NFE.}
    \label{fig:samples_afhq}
\end{figure*}

\begin{figure*}[!tp]
    \centering
    \includegraphics[width=\textwidth]{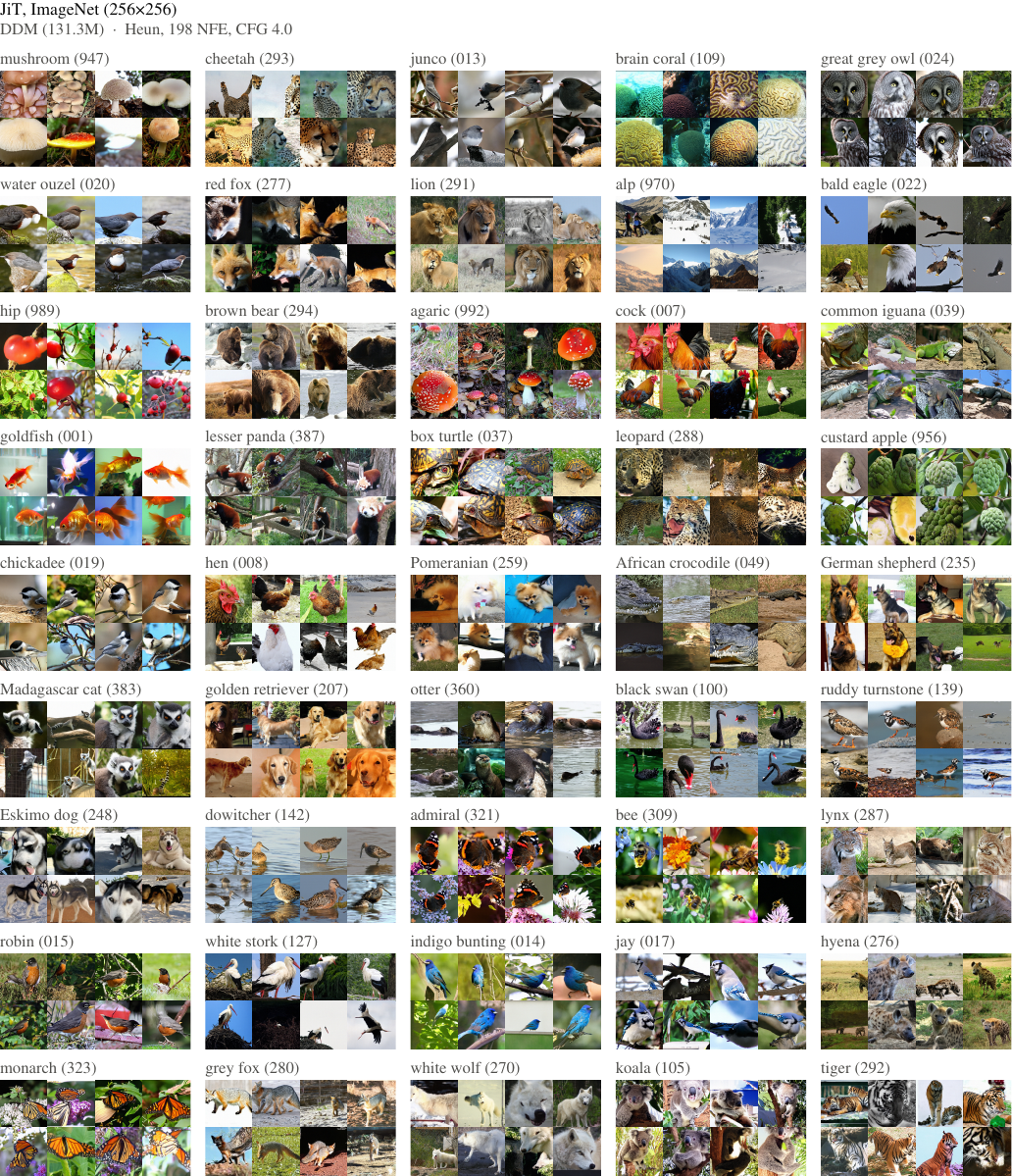}
    \caption{Uncurated samples of the \method's JiT-B/16 (131.3M) student on ImageNet at $256\times256$, with Heun's solver at 198 NFE and guidance 4.0 for $t\in[0.1,1]$.}
    \label{fig:jit_pool}
\end{figure*}

Figs.~\ref{fig:samples_small}--\ref{fig:samples_afhq} show uncurated samples of the 4$\times$ smaller \method\ students, and Fig.~\ref{fig:jit_pool} uncurated samples of the \method\ JiT-B/16 student for 45 ImageNet classes.

\section{Student and teacher trajectories from the same noise}\label{appendix:velocity_prediction}
\begin{figure*}[!tp]
    \centering
    \includegraphics[width=\textwidth]{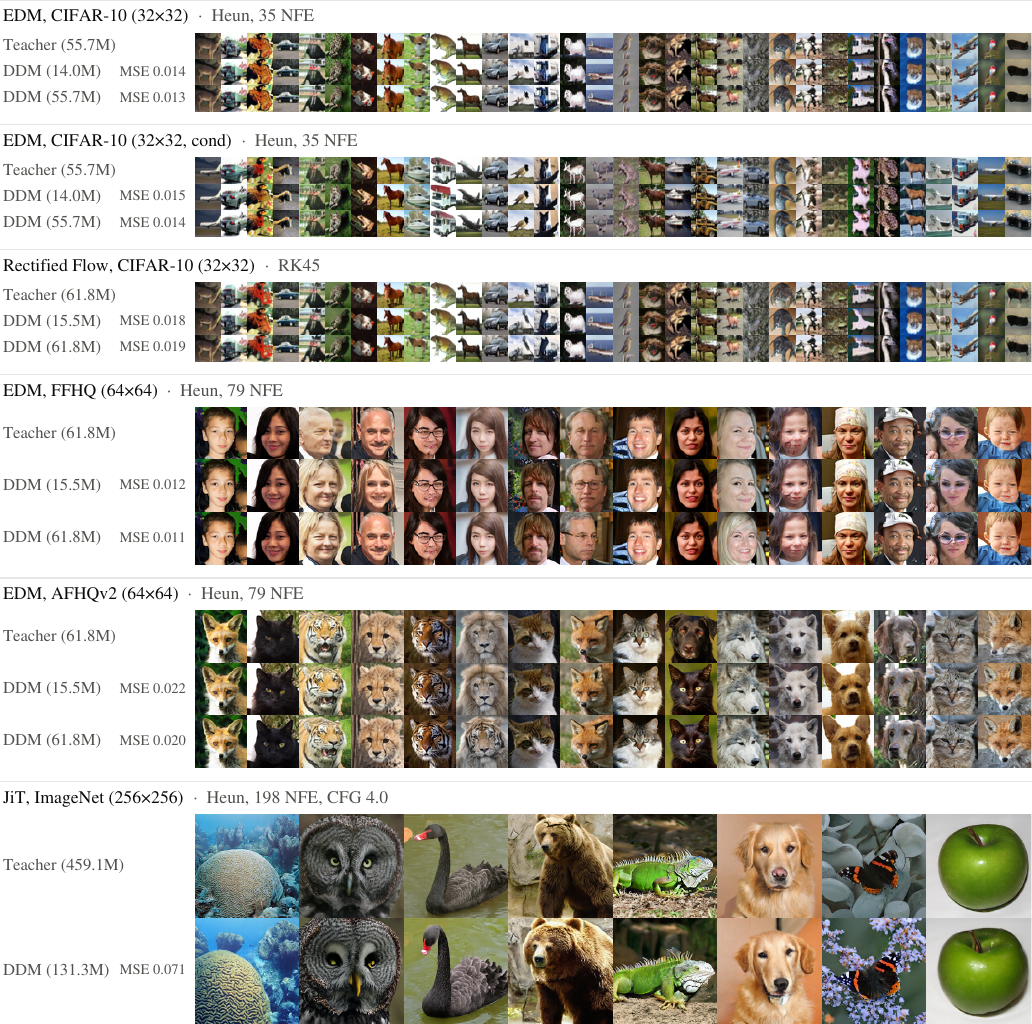}
    \caption{Teacher and \method\ students sampled from the same initial noise with the teacher's sampler. The students largely reproduce their teacher's noise-to-image map. MSE gives the mean squared difference to the teacher's images over 1024 samples per model.}
    \label{fig:samples_pairs}
\end{figure*}

\begin{figure*}[!tp]
    \centering
    \includegraphics[width=\textwidth]{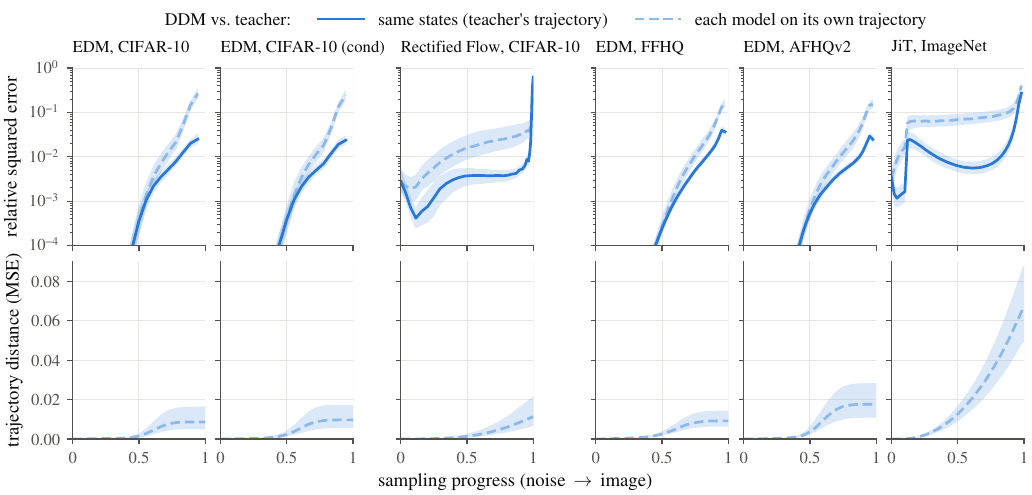}
    \caption{Teacher and \method\ students sampled from the same initial noise with the teacher's sampler, for the smaller student of each setting of Fig.~\ref{fig:samples_pairs} (1024 samples each). Top, relative squared error $\|v^S-v^T\|^2/\|v^T\|^2$ at the start of every step, on the teacher's trajectory (solid) and with each model on its own trajectory (dashed). Bottom, squared distance between the states of the two trajectories per pixel value. Lines show medians and bands the interquartile range over the 1024 samples.}
    \label{fig:velocity_prediction_comp}
\end{figure*}

Fig.~\ref{fig:samples_pairs} samples each teacher and its \method\ students from the same initial noise with the teacher's sampler, and Fig.~\ref{fig:velocity_prediction_comp} follows the two trajectories of 1024 such pairs. On the teacher's states, the median relative squared velocity error of the student stays below $10^{-4}$ over the first 40\% of the EDM grids and grows towards the data, to 2.4--3.7\% at the last step. For JiT it stays below 2.5\% up to $t=0.9$, and for Rectified Flow between 0.04\% and 1.3\% up to $t=0.98$, beyond which no training grid reaches, since the last step of \method\ lies at $(\omega-1)/\omega\le0.98$. When each model follows its own trajectory, the states of the two models differ slightly, and near the data, where the velocity field is steep, this difference increases the velocity error. The final images differ by a mean squared error of 0.011 to 0.022, and 0.071 for JiT, on pixel values in $[-1,1]$.

\section{Additional ablations}\label{appendix:ablation_rf}\label{appendix:random_positions}
\begin{figure}[!tp]
    \centering
    \includegraphics[width=\linewidth]{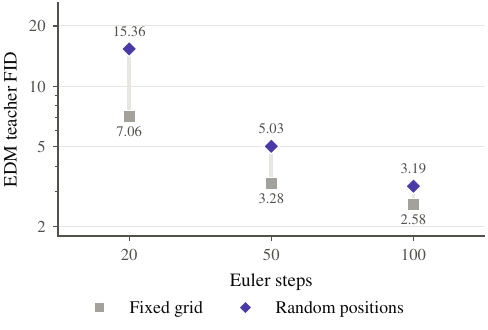}
    \caption{FID of the EDM CIFAR-10 teacher sampled with Euler on its fixed grid and with per-sample random positions (50K samples).}
    \label{fig:random_positions}
\end{figure}

\textbf{Pool size.} With a pool of $M=10b$ samples, \method\ attains 4.66 FID on EDM, against 4.76 for $M=3b$ and 5.17 for $M=b$, but 5.53 on Rectified Flow, against 5.23 and 5.44. A pool larger than $M=3b$ thus brings no consistent gain.

\textbf{Random positions.} Fig.~\ref{fig:random_positions} shows what the large steps of random positions (Fig.~\ref{fig:coverage}c) cost on the curved ODE of the EDM teacher. Random positions raise the FID of the teacher at every number of steps, from 7.06 to 15.36 at 20 steps and from 2.58 to 3.19 at 100 steps, as consecutive positions may lie far apart even on fine grids. In training with a pool and unrolling with Heun's solver, however, the student is less sensitive to these steps, and random positions perform comparably to \method\ (Tab.~\ref{tab:randomization}).

\textbf{Fixed grid on Rectified Flow.} Sampled with Euler's solver on its own 30 training levels, the fixed-grid student with a pool of $M=3b$ attains 7.75 FID, on par with the 7.89 of \method\ at the same sampler. With RK45, which evaluates it between these levels, it attains 15.66 FID at 644 NFE, against 5.23 at 278 NFE for \method. Independence thus renders the fixed-grid student accurate at its 30 supervised noise levels only, and a continuous-time sampler, which evaluates the student at arbitrary noise levels, requires the higher density of noise levels that \method\ provides.

\section{Comparison with teacher supervision on real images}\label{appendix:additional_comparisons}
Tab.~\ref{tab:dockhorn} compares \method\ under the protocol of \citet{dockhorn2023distilling}, who train a 19.8M student on real CIFAR-10 images with the teacher's denoiser as the target and report the FID on 5K samples. With the same architecture and without stored images, \method\ attains 7.98 against 7.63, and 7.52 with a slightly larger student of 25.6M parameters. Without any real image, \method\ thus comes within 0.35 FID of teacher supervision on real data at the same architecture, and surpasses it with a slightly larger student.

\begin{table}[H]
  \centering
  \footnotesize
  \setlength{\tabcolsep}{4pt}
  \caption{\method\ vs. teacher supervision on real images under the protocol of \citet{dockhorn2023distilling}, FID on 5K samples. We evaluate the teacher under this protocol. \citet{dockhorn2023distilling} do not report the NFE of their student. Column ``Data'' denotes stored images with ``\dataR '' for real and \dataN\ for none.}
  \label{tab:dockhorn}
  \begin{tabular}{clllrl}
\toprule
Data & Model & Params & Sampler & NFE & FID$\downarrow$ \\
\midrule
\addlinespace
\multicolumn{6}{l}{\textbf{\boldmath EDM, CIFAR-10 ($32\times32$) $\cdot$ FID-5K, protocol of Dockhorn et al.}} \\
\dataR & Teacher~\citep{karras2022elucidating} & 55.7M & Heun & 35 & 6.58 \\
\dataR & \citet{dockhorn2023distilling} & 19.8M & Heun & -- & 7.63 \\
\cmidrule{1-6}
\dataN &  & 19.8M & Heun & 35 & 7.98 \\
\dataN & \multirow{-2}{*}{\method\ (ours)} & 25.6M & Heun & 35 & 7.52 \\
\bottomrule
\end{tabular}

\end{table}

\section{Memorization}\label{appendix:memorization}
We test whether the students of Tab.~\ref{tab:students_vs_teachers} that are $4\times$ smaller than their teachers, trained with \method\ and with standard training, and their EDM teachers reproduce training images. For each of 10K samples per model, or 5K for the teachers at $64\times64$, we find its two nearest training images in pixel space, counting a training image and its horizontal flip as one image, and consider the sample memorized when its distance to the nearest image is below a third of its distance to the second nearest~\citep{yoon2023diffusion}. No sample is memorized on CIFAR-10, class-conditional CIFAR-10, FFHQ and AFHQv2, for \method\ and the teachers and for standard training after 40K, 200K and 400K iterations. At 400K iterations, the smallest ratio of the two distances is 0.67 to 0.78 for \method\ and 0.58 to 0.75 for standard training, and 0.63 to 0.82 for the teachers, far above $1/3$. The test detects near-duplicates among real images, as it flags 19 of 10K CIFAR-10 test images with respect to the training set, and 6 and 11 of 2K training images of CIFAR-10 and AFHQv2 with respect to the remaining training images. The median root-mean-square pixel distance of a sample to its nearest training image is 0.179 for \method\ and 0.168 for standard training on CIFAR-10, 0.192 and 0.183 on FFHQ, and 0.194 and 0.183 on AFHQv2, against 0.166, 0.184 and 0.192 between a training image and its nearest other training image. The samples of \method\ thus lie slightly farther from the training images than those of standard training.

\section{Limitations, scope and impact}\label{appendix:limitations}\label{appendix:ethics}
\textbf{Sample quality.} \method\ transfers the generative behavior of a pretrained teacher instead of learning directly from the original data distribution. Consequently, its achievable quality is constrained by the teacher's learned distribution, and its students may inherit the teacher's biases and errors. Standard training, which learns directly from the stored images, attains a lower FID than \method\ at the size of the teacher, by 0.75 to 1.92, and for the smaller students on FFHQ and AFHQv2, by 0.99 and 0.23 (Tab.~\ref{tab:students_vs_teachers}). Moreover, the student is supervised on the states visited by the teacher but follows its own trajectories at inference time. This discrepancy may lead to accumulated errors, particularly near the end of sampling, as illustrated by our trajectory analysis in App.~\ref{appendix:velocity_prediction}.

\textbf{Computational and storage costs.} The computational cost of \method\ depends on the space in which the models operate. In pixel space, the evaluation of the teacher at every iteration lowers the training throughput to 0.42 to 0.73$\times$ that of standard training, and to 0.20$\times$ for the JiT student. In latent space, where standard training encodes full-resolution images at every iteration, \method\ trains 1.40$\times$ faster on CelebA-HQ and FFHQ at $256\times256$ (App.~\ref{appendix:cost}). The storage advantage grows with the size of the training set, as the weights of the teacher occupy more space than the images of CIFAR-10 and AFHQv2, but 137$\times$ less than those of ImageNet at $256\times256$ (Tab.~\ref{tab:storage}).

\textbf{Scope of the experiments.} Although \method\ trains students of any architecture, our experiments establish the feasibility of cross-architecture transfer (Tab.~\ref{tab:cross_arch}) without aiming to optimize generation performance. The comparisons with modified step reduction methods (Tab.~\ref{tab:small_students}) merely evaluate their applicability under random initialization and the absence of stored images.

\textbf{Broader impact.} \method\ trains new models without storing any training data, from the weights of a pretrained teacher alone. It thereby removes the need for large dataset storage and allows new models to be trained where the training data cannot be shared, for instance for privacy or licensing reasons. The test of App.~\ref{appendix:memorization} finds no reproduced training image among the samples of our students on CIFAR-10, FFHQ and AFHQv2, but the student may still inherit vulnerabilities of its teacher, such as membership inference or data extraction from the learned velocity field. High-quality synthetic media may contribute to the proliferation of deepfakes and misinformation, and protection measures are available to mitigate this risk.

\end{document}